\documentclass[10pt,journal,compsoc]{IEEEtran}
\ifCLASSOPTIONcompsoc
  \usepackage[nocompress]{cite}
\else
  \usepackage{cite}
\fi
\ifCLASSINFOpdf
\else
\fi

\usepackage{epsfig}
\usepackage{graphicx}
\usepackage{amsmath}
\usepackage{amssymb}
\usepackage{enumitem}

\usepackage[pagebackref=true,breaklinks=true,letterpaper=true,colorlinks,bookmarks=false,urlcolor=red,citecolor=cyan]{hyperref}
\usepackage{gensymb,graphics,subfigure,inputenc}
\usepackage[linesnumbered,algo2e,boxed]{algorithm2e}
\graphicspath{{Figure/}}
\usepackage[figuresright]{rotating}
\usepackage{amsmath,amssymb,textcomp,subfigure,multirow,upgreek}
\usepackage{booktabs}
\usepackage{color,mdwlist}

\usepackage{graphicx,fontawesome}
\graphicspath{{Figures/}}

\def\ci#1{\textcircled{\resizebox{.4em}{!}{#1}}}

\usepackage{caption}
\usepackage{enumitem}
\setitemize{noitemsep,topsep=0pt,parsep=0pt,partopsep=0pt}

\begin{document}
%
% paper title
% Titles are generally capitalized except for words such as a, an, and, as,
% at, but, by, for, in, nor, of, on, or, the, to and up, which are usually
% not capitalized unless they are the first or last word of the title.
% Linebreaks \\ can be used within to get better formatting as desired.
% Do not put math or special symbols in the title.
\title{Multi-Channel Attention Selection GANs for Guided Image-to-Image Translation}
\author{Hao~Tang,
% 	Dan~Xu,
%  	Yan~Yan,
    % Jason J. Corso,
 	Philip H.S. Torr,
	Nicu~Sebe
	\IEEEcompsocitemizethanks{
	    \IEEEcompsocthanksitem Hao Tang is with the Department of Information Technology and Electrical Engineering, ETH Zurich,  Zurich 8092, Switzerland. E-mail: hao.tang@vision.ee.ethz.ch \protect
        % \IEEEcompsocthanksitem Hong Liu is with the Shenzhen Graduate School, Peking University, Shenzhen 518055, China. \protect
	    \IEEEcompsocthanksitem Philip H.S. Torr is with the Department of Engineering Science, University of Oxford, Oxford OX1 2JD, United Kingdom. \protect
        % \IEEEcompsocthanksitem Yan Yan is with the Department of Computer Science,  Texas State University, San Marcos 78666, USA. \protect
        % \IEEEcompsocthanksitem Jason J. Corso is with the  Department of Electrical Engineering and Computer Science, University of Michigan, Ann Arbor 48109, USA. \protect
        \IEEEcompsocthanksitem Nicu Sebe is with the Department of Information Engineering and Computer Science (DISI), University of Trento, Trento 38123, Italy. \protect
        }% <-this % stops an unwanted space
}

% note the % following the last \IEEEmembership and also \thanks - 
% these prevent an unwanted space from occurring between the last author name
% and the end of the author line. i.e., if you had this:
% 
% \author{....lastname \thanks{...} \thanks{...} }
%                     ^------------^------------^----Do not want these spaces!
%
% a space would be appended to the last name and could cause every name on that
% line to be shifted left slightly. This is one of those "LaTeX things". For
% instance, "\textbf{A} \textbf{B}" will typeset as "A B" not "AB". To get
% "AB" then you have to do: "\textbf{A}\textbf{B}"
% \thanks is no different in this regard, so shield the last } of each \thanks
% that ends a line with a % and do not let a space in before the next \thanks.
% Spaces after \IEEEmembership other than the last one are OK (and needed) as
% you are supposed to have spaces between the names. For what it is worth,
% this is a minor point as most people would not even notice if the said evil
% space somehow managed to creep in.

% The paper headers
\markboth{IEEE Transactions on Pattern Analysis and Machine Intelligence}%
{Shell \MakeLowercase{\textit{et al.}}: Bare Demo of IEEEtran.cls for Computer Society Journals}
% The only time the second header will appear is for the odd numbered pages
% after the title page when using the twoside option.
% 
% *** Note that you probably will NOT want to include the author's ***
% *** name in the headers of peer review papers.                   ***
% You can use \ifCLASSOPTIONpeerreview for conditional compilation here if
% you desire.

% The publisher's ID mark at the bottom of the page is less important with
% Computer Society journal papers as those publications place the marks
% outside of the main text columns and, therefore, unlike regular IEEE
% journals, the available text space is not reduced by their presence.
% If you want to put a publisher's ID mark on the page you can do it like
% this:
%\IEEEpubid{0000--0000/00\$00.00~\copyright~2015 IEEE}
% or like this to get the Computer Society new two part style.
%\IEEEpubid{\makebox[\columnwidth]{\hfill 0000--0000/00/\$00.00~\copyright~2015 IEEE}%
%\hspace{\columnsep}\makebox[\columnwidth]{Published by the IEEE Computer Society\hfill}}
% Remember, if you use this you must call \IEEEpubidadjcol in the second
% column for its text to clear the IEEEpubid mark (Computer Society jorunal
% papers don't need this extra clearance.)

% use for special paper notices
%\IEEEspecialpapernotice{(Invited Paper)}

% for Computer Society papers, we must declare the abstract and index terms
% PRIOR to the title within the \IEEEtitleabstractindextext IEEEtran
% command as these need to go into the title area created by \maketitle.
% As a general rule, do not put math, special symbols or citations
% in the abstract or keywords.
\IEEEtitleabstractindextext{%
%\begin{abstract}
%The abstract goes here.
%\end{abstract}
\begin{abstract}
We propose a novel model named Multi-Channel Attention Selection Generative Adversarial Network (SelectionGAN) for guided image-to-image translation, where we translate an input image into another while respecting an external semantic guidance. The proposed SelectionGAN explicitly utilizes the semantic guidance information and consists of two stages. In the first stage, the input image and the conditional semantic guidance are fed into a cycled semantic-guided generation network to produce initial coarse results. In the second stage, we refine the initial results by using the proposed multi-scale spatial pooling \& channel selection module and the multi-channel attention selection module. Moreover, uncertainty maps automatically learned from attention maps are used to guide the pixel loss for better network optimization. Exhaustive experiments on four challenging guided image-to-image translation tasks (face, hand, body, and street view) demonstrate that our SelectionGAN is able to generate significantly better results than the state-of-the-art methods. Meanwhile, the proposed framework and modules are unified solutions and can be applied to solve other generation tasks such as semantic image synthesis. The code is available at~\url{https://github.com/Ha0Tang/SelectionGAN}.
\end{abstract}

% Note that keywords are not normally used for peerreview papers.
}

% make the title area
\maketitle

% To allow for easy dual compilation without having to reenter the
% abstract/keywords data, the \IEEEtitleabstractindextext text will
% not be used in maketitle, but will appear (i.e., to be "transported")
% here as \IEEEdisplaynontitleabstractindextext when the compsoc 
% or transmag modes are not selected <OR> if conference mode is selected 
% - because all conference papers position the abstract like regular
% papers do.
\IEEEdisplaynontitleabstractindextext
% \IEEEdisplaynontitleabstractindextext has no effect when using
% compsoc or transmag under a non-conference mode.

% For peer review papers, you can put extra information on the cover
% page as needed:
% \ifCLASSOPTIONpeerreview
% \begin{center} \bfseries EDICS Category: 3-BBND \end{center}
% \fi
%
% For peerreview papers, this IEEEtran command inserts a page break and
% creates the second title. It will be ignored for other modes.
\IEEEpeerreviewmaketitle

% Computer Society journal (but not conference!) papers do something unusual
% with the very first section heading (almost always called "Introduction").
% They place it ABOVE the main text! IEEEtran.cls does not automatically do
% this for you, but you can achieve this effect with the provided
% \IEEEraisesectionheading{} command. Note the need to keep any \label that
% is to refer to the section immediately after \section in the above as
% \IEEEraisesectionheading puts \section within a raised box.

% The very first letter is a 2 line initial drop letter followed
% by the rest of the first word in caps (small caps for compsoc).
% 
% form to use if the first word consists of a single letter:
% \IEEEPARstart{A}{demo} file is ....
% 
% form to use if you need the single drop letter followed by
% normal text (unknown if ever used by the IEEE):
% \IEEEPARstart{A}{}demo file is ....
% 
% Some journals put the first two words in caps:
% \IEEEPARstart{T}{his demo} file is ....
% 
% Here we have the typical use of a "T" for an initial drop letter
% and "HIS" in caps to complete the first word.

\IEEEraisesectionheading{\section{Introduction}
	\label{sec:introduction}}

\begin{figure*}[!t] \small
	\centering
	\includegraphics[width=0.85\linewidth]{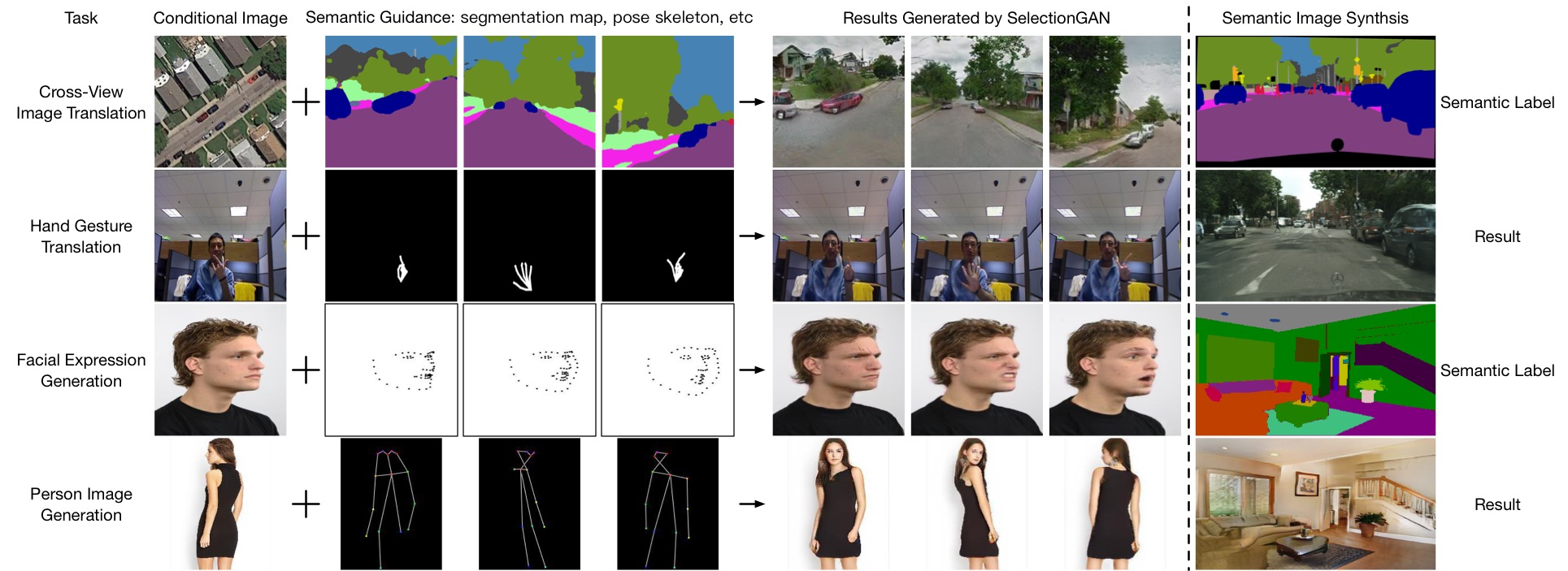}
	\caption{SelectionGAN's capabilities: (\textit{left}) Guided image-to-image translation (including cross-view image translation, hand gesture translation, facial expression generation, and person image generation): synthesizing images from a single input image as well as semantic guidance (e.g., segmentation map, hand skeleton, facial landmark, and human pose skeleton). (\textit{right}) Semantic image synthesis: SelectionGAN simultaneously produces realistic images while respecting the spatial semantic layout for both outdoor and indoor scenes.}
	\label{fig:motivation}
	\vspace{-0.2cm}
\end{figure*}

\IEEEPARstart{G}{uided} image-to-image translation is aiming at synthesizing new images from an input image and several external semantic guidance, as shown in Fig.~\ref{fig:motivation}.  
This task received a lot interest especially from the computer vision community, and has been widely investigated in recent years. 
Due to different forms of semantic guidance, e.g., segmentation maps, hand skeletons, facial landmarks, etc., most existing methods are tailored toward specific applications, i.e., they need to specifically design the network architectures and training objectives according to different generation tasks.
For example, Ma et al. propose PG2~\cite{ma2017pose}, which is a two-stage framework and uses the pose mask loss for generating person images based on an image of that person and human pose keypoints.
Tang et al. propose GestureGAN~\cite{tang2018gesturegan}, which is a forward-backward consistency architecture and adopt a novel color loss to generate novel hand gesture images based on the input image and conditional hand skeletons.
Wang et al. propose the few-shot Vid2Vid framework~\cite{wang2019few}, which uses the carefully designed weight generation module to synthesize videos that realistically reflect the style of the input image and the layout of conditional segmentation maps.

Different from previous works in guided image-to-image translation, in this paper, we focus on developing a framework that is application-independent.
This makes our framework and modules more widely applicable to many generation tasks with different forms of semantic guidance.
To tackle this challenging problem, AlBahar and Huang~\cite{albahar2019guided} recently propose a bi-directional feature transformation to better utilize the constraints of the semantic guidance. 
Although this approach performs an interesting exploration, we observe unsatisfactory aspects mainly in the generated image layout and content details, which are due to three different reasons. 
First, since it is always costly to obtain manually annotated semantic guidance, the semantic guidance is usually produced from pre-trained models trained on other large-scale datasets, e.g., pose skeletons are extracted using OpenPose \cite{amos2016openface} and segmentation maps are extracted using \cite{lin2017refinenet,zhou2017scene}, leading to insufficiently accurate predictions for all the pixels, and thus misguiding the image generation process. 
Second, we argue that the translation with a single phase generation network is not able to capture the complex image structural relationships between the source and target domains, especially when source and target domains only have little or even no overlap, e.g., person image generation \cite{ma2017pose,tang2020xinggan}, and cross-view image translation \cite{tang2019multi,regmi2018cross}. 
Third, a three-channel generation space may not be suitable enough for learning a good mapping for this complex synthesis problem. 
Given these problems, could we enlarge the generation space and learn an automatic selection mechanism to synthesize more fine-grained generation results?

\begin{figure*} [!t] \small
	\centering
	\includegraphics[width=0.85\linewidth]{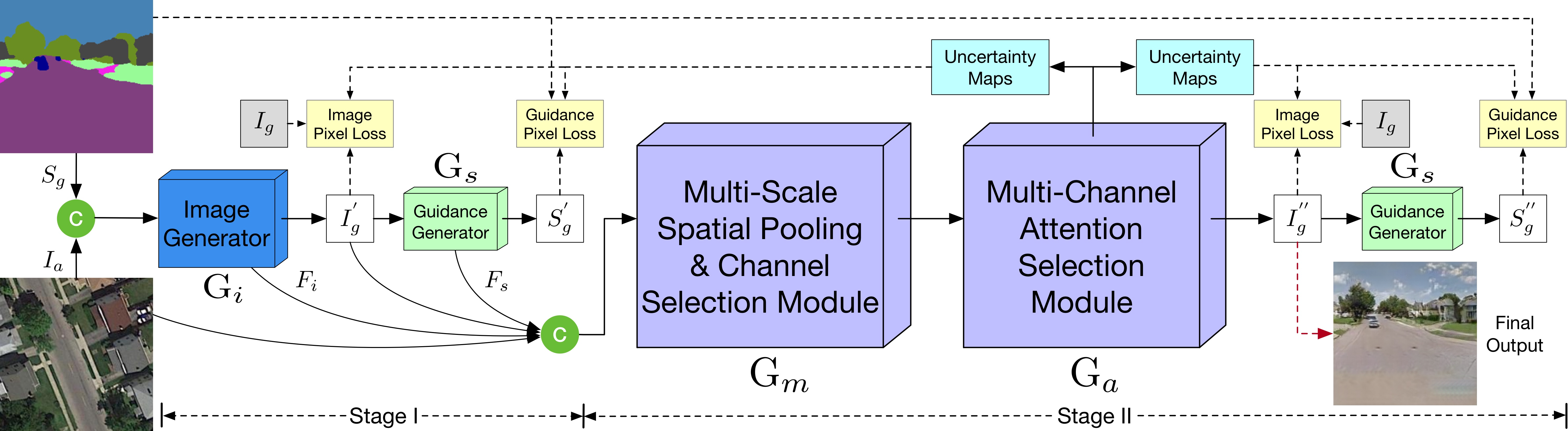}
	\caption{Overview of the proposed SelectionGAN. Stage I presents a cycled semantic-guided generation sub-network which accepts both the input image $I_a$ and the conditional semantic guidance $S_g$, and simultaneously synthesizes the target image $I_g^{'}$ and reconstructs the semantic guidance $S_g^{'}$. 
	Stage II takes the input image $I_a$, the coarse prediction $I_g^{'}$, and the learned deep features ($F_i$ and $F_s$) from stage I, and performs a fine-grained generation using the proposed multi-scale spatial pooling \& channel selection and the multi-channel attention selection modules. $\textcircled{c}$ denotes channel-wise concatenation.}
	\label{fig:framework}
	\vspace{-0.2cm}
\end{figure*}

Based on these observations we propose a novel Multi-Channel Attention Selection Generative Adversarial Network (SelectionGAN), which contains two generation stages. 
The overall framework of SelectionGAN is shown in Fig.~\ref{fig:framework}.
In the first stage, we learn a cycled image-guidance generation sub-network, which accepts a pair consisting of an image and the conditional semantic guidance, and generates target images, which are further fed into a semantic guidance generation network to reconstruct the input semantic guidance. This cycled guidance generation adds stronger supervision between the image and guidance domains, facilitating the optimization of the network. 

The coarse outputs from the first generation network, including the input image, together with the deep feature maps from the last layer, are input into the second stage networks.
We first employ the proposed multi-scale spatial pooling \& channel selection module to enhance the multi-scale features in both spatial and channel dimensions.
Next, several intermediate outputs are produced, and simultaneously we learn a set of multi-channel attention maps with the same number as the intermediate generations. These attention maps are used to spatially select from the intermediate generations, and are combined to synthesize a final output.
Finally, to overcome the inaccurate semantic guidance issue, the multi-channel attention maps are further used to generate uncertainty maps to guide the reconstruction loss. 
Through extensive experimental evaluations, we demonstrate that SelectionGAN produces remarkably better results than the existing baselines on four different guided image-to-image translation tasks, i.e., segmentation map guided cross-view image translation, hand skeleton guided gesture-to-gesture translation, facial landmark guided expression-to-expression translation, and pose guided person image generation. 
Moreover, the proposed framework and modules can be applied to other generation tasks such as semantic image synthesis.

Overall, the contributions of this paper are as follows:
\begin{itemize}[leftmargin=*]
	\item A novel Multi-Channel Attention Selection GAN (SelectionGAN) for guided image-to-image translation task is presented. It explores cascaded semantic guidance with a coarse-to-fine inference, and aims at producing a more detailed synthesis from richer and more diverse multiple intermediate generations.
	\item A novel multi-scale spatial pooling \& channel selection module is proposed, which is utilized to automatically enhance the multi-scale feature representation in both spatial and channel dimensions. 
	\item A novel multi-channel attention selection module is proposed, which is utilized to attentively select interested intermediate generations and is able to significantly boost the quality of the final output. The multi-channel attention module also effectively learns uncertainty maps to guide the pixel loss for more robust optimization. 
	\item Extensive experiments clearly demonstrate the effectiveness of the proposed SelectionGAN, and show state-of-the-art results on four guided image-to-image translation (including face, hand, body, and street view) tasks. Moreover, we show the proposed SelectionGAN is effective on other generation tasks such as semantic image synthesis.
\end{itemize}

Part of the material presented here appeared in \cite{tang2019multi}. The current paper extends \cite{tang2019multi} in several ways.
(1) We present a more detailed analysis of related works by including recently published works dealing with guided image-to-image translation. 
(2) We propose a novel module, i.e., multi-scale channel selection, to automatically enhance the multi-scale feature representation in the feature channel dimension. 
Equipped with this new module, our SelectionGAN proposed in \cite{tang2019multi} is upgraded to SelectionGAN++.
(3) We extent the proposed framework to a more robust and general framework for handling different guided image-to-image translation tasks. 
(4) We extend the quantitative and qualitative experiments by comparing our SelectionGAN and SelectionGAN++ with the very recent works on four guided image-to-image translation tasks and one semantic image synthesis task with 11 public datasets.

\section{Related Work}
\label{sec:related}

\noindent\textbf{Generative Adversarial Networks (GANs)}~\cite{goodfellow2014generative} have shown the capability of generating high-quality images \cite{karras2019style}.
A vanilla GAN model~\cite{goodfellow2014generative} has two important components: a generator $G$ and a discriminator $D$.
The goal of $G$ is to generate photo-realistic images from a noise vector, while~$D$ is trying to distinguish between a real image and the image generated by $G$.
Although it is successfully used in generating images of high visual fidelity, there are still some challenges, i.e., how to generate images in a conditional setting. 
To generate domain-specific images, Conditional GANs (CGANs)~\cite{mirza2014conditional} have been proposed.
One specific application of CGANs is image-to-image translation~\cite{isola2017image}.

\noindent\textbf{Image-to-Image Translation} frameworks learn a parametric mapping between inputs and outputs.
For example, Isola et al.~\cite{isola2017image} propose Pix2pix, which is a supervised model and uses a CGAN to learn a translation function from input to output image domains.
Based on Pix2pix, Wang et al.~\cite{wang2018high} propose Pix2pixHD, which can turn semantic maps into photo-realistic images.

Our work builds upon the recent advances in image-to-image translation, i.e., Pix2pix, and aims to extend it to a broader set of guided image-to-image translation problem, which provides users with more input.
Moreover, the proposed multi-scale spatial pooling \& channel selection and the multi-channel attention selection modules are network-agnostic and can be plugged into any existing GAN-based generation architectures.

\noindent\textbf{Guided Image-to-Image Translation} is a variant of image-to-image translation problem aimed at translating an input image to a target image while respecting certain constrains specified by some external guidance, such as class labels~\cite{choi2018stargan,tang2019dual,tang2019expression}, text descriptions \cite{xu2022predict,liu2020open,tao2022df}, human keypoint/skeleton~\cite{tang2018gesturegan,ma2017pose,tang2020unified,tang2021total,tang2020bipartite,tang2020xinggan}, segmentation maps~\cite{tang2019multi,regmi2018cross,wang2019few,tang2020dual,liu2020exocentric,tang2019local,wu2022cross,wu2022crossview,tang2022local,ren2021cascaded,tang2021layout,liu2021cross}, and reference images~\cite{wang2019example,albahar2019guided}.
Given that different generation tasks need different guidance information, existing works are tailored to a specific application, i.e., with specifically designed network architectures and training objectives.
For example, Ma et al. propose PG2~\cite{ma2017pose}, which is a two-stage framework and uses the pose mask loss for generating person images based on an image of that person and human pose keypoints.
Tang et al. propose GestureGAN~\cite{tang2018gesturegan}, which is a forward-backward consistency architecture and adopt the proposed color loss to generate novel hand gesture images based on the input image and conditional hand skeletons.
Wang et al. propose the few-shot Vid2Vid framework~\cite{wang2019few}, which uses a carefully designed weight generation module to synthesize videos that realistically reflect the style of the input image and the layout of conditional segmentation maps.

Compared to existing works in guided image-to-image translation, we develop a unified and robust framework that is application-independent. 
In this way, the proposed framework can be widely applied to many generation tasks with different forms of guidance such as scene segmentation maps, hand skeletons, facial landmarks, and human body skeleton (see Fig.~\ref{fig:motivation}).

\noindent\textbf{Attention Learning in Image-to-Image Translation.}
Attention learning has been extensively exploited in computer vision and natural language processing, e.g., \cite{xu2018structured,vaswani2017attention,ding2020lanet,duan2019cascade,shi2022charformer,yang2022continual}.
To improve the image generation performance, the attention mechanism has also been recently investigated in GANs such as \cite{tang2019attentiongan,kim2019u,zhang2018self,tang2019attention}. 
For example, Zhang et al. propose SAGAN \cite{zhang2018self}, which introduces a self-attention
mechanism into convolutional GANs to help with
modeling long range, multi-level dependencies across image regions.

Unlike existing attention methods, we aim at a more effective network design and propose a novel SelectionGAN, which allows to automatically select from multiple diverse and rich intermediate generations, and thus significantly improving the generation quality.
To the best of our knowledge, our model is the first attempt to incorporate a multi-channel attention selection module within a GAN framework for image-to-image translation tasks.
\section{SelectionGAN}
\label{sec:method}

\begin{figure*}[!t] \small
	\centering
	\includegraphics[width=0.85\linewidth]{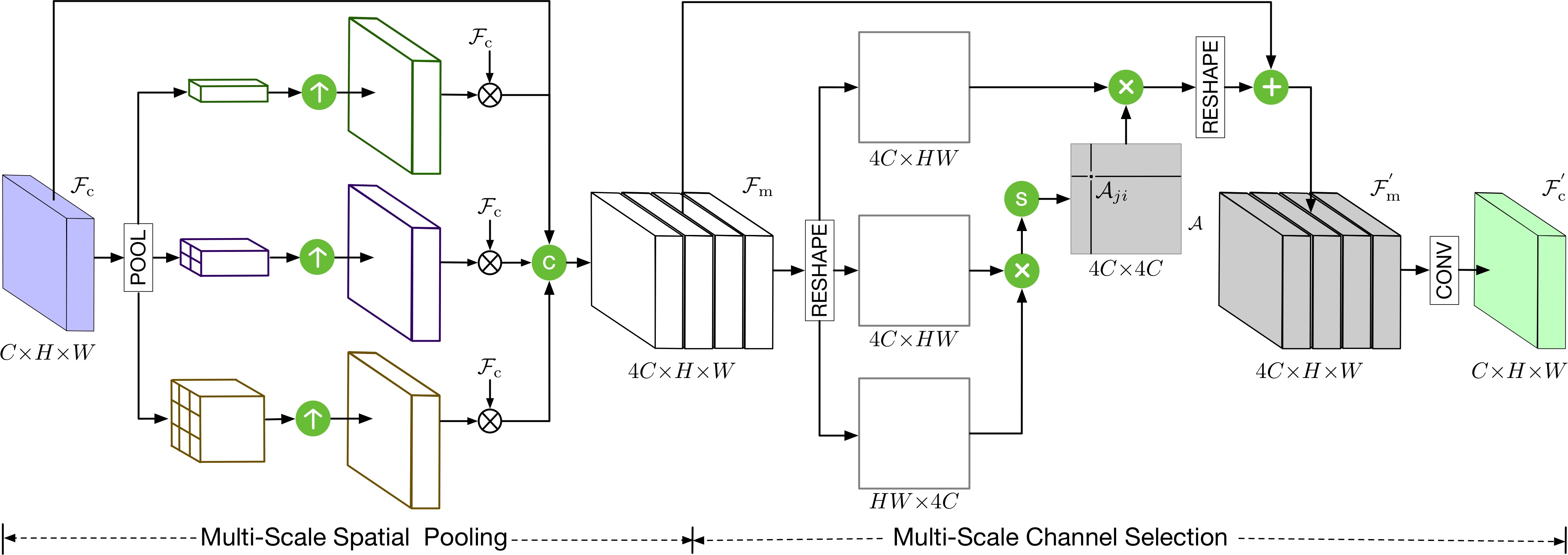}
	\caption{Overview of the proposed multi-scale spatial pooling \& channel selection module. The multi-scale spatial pooling pools features from different receptive fields in order to have a better generation of image details. The multi-scale channel selection aims at automatically emphasizing interdependent channel maps by integrating associated features among all multi-scale channel maps to improve deep feature representation. $\oplus$, $\otimes$, $\textcircled{c}$, $\textcircled{s}$ and \protect\ci{$\uparrow$} denote element-wise addition, element-wise multiplication, channel-wise concatenation, softmax, and up-sampling operation, respectively.}
	\label{fig:channel}
	\vspace{-0.2cm}
\end{figure*}

In this section we present the details of the proposed multi-channel attention selection GAN. An illustration of the overall network structure is depicted in Fig.~\ref{fig:framework}. 
In the first stage, we present a cascaded semantic-guided generation sub-network, which utilizes the input image and the conditional semantic guidance as inputs, and generate the target images while respecting the semantic guidance.

These generated images are further input into a semantic guidance generator to recover the input guidance forming a generation cycle. In the second stage, the coarse synthesis and the deep features from the first stage are combined, and then are passed to the proposed multi-scale spatial pooling \& channel selection module to model the long-range multi-scale dependencies between each channel of feature representations. Thus the enhanced feature maps are fed to the proposed multi-channel attention selection module, which aims at producing more fine-grained synthesis from a larger generation space and also at generating uncertainty maps to jointly guide multiple optimization losses.

\subsection{Cascade Semantic-Guided Generation}

\noindent \textbf{Semantic-Guided Generation.}
We target to translate an input image to another while respecting the semantic guidance. There are many strategies to incorporate the additional semantic guidance into the image-to-image translation model \cite{albahar2019guided,tang2020xinggan} and the most straight forward scheme is input concatenation.
Specifically, as shown in Fig.~\ref{fig:framework}, we concatenate the input image $I_a$ and the semantic guidance $S_g$, and feed them into the image generator $G_i$ and synthesize the target image $I_g^{'}$ as $I_g^{'} {=} G_i(I_a, S_g)$.
In this way, the semantic guidance provides stronger supervision to guide the image-to-image translation in the deep network.

\noindent \textbf{Semantic-Guided Cycle.}
Existing guided image-to-image translation methods \cite{ma2017pose,siarohin2018deformable,albahar2019guided} only use semantic guidance as input to guide the image generation process, which actually provide a weak guidance.
Different from theirs, we apply the semantic guidance not only as input but also as part of the network's output.
Specifically, as shown in Fig.~\ref{fig:framework}, we propose a cycled semantic guidance generation network to benefit more the semantic guidance information in learning jointly. The conditional semantic guidance $S_g$ together with the input image $I_a$ are input into the image generator $G_i$, and produce the synthesized image $I_g^{'}$. Then $I_g^{'}$ is further fed into the semantic guidance generator $G_s$ which reconstructs a new semantic guidance $S_g^{'}$. 
We can formalize the process as $S_g^{'} {=} G_s(I_g^{'}){=}G_s(G_i(I_a, S_g))$. 
Then the optimization objective is to make $S_g^{'}$ as close as possible to $S_g$, which naturally forms a semantic guidance generation cycle, i.e.,~$[I_a, S_g] \stackrel{G_i} \rightarrow I_g^{'} \stackrel{G_s} \rightarrow S_g^{'}  {\approx} S_g$. 
The two generators are explicitly connected by the ground-truth semantic guidance, which in this way provides extra constraints on the generators to better learn the semantic structure consistency. 
We observe that the simultaneous generation of both the images and the semantic guidance improves the generation performance in our experiments section. 

\noindent \textbf{Cascade Generation.} Due to the complexity of the tasks such as in pose guided person image generation \cite{ma2017pose,tang2020bipartite,tang2020xinggan}, input and output domains usually have little overlap, which apparently leads to ambiguity issues in the generation process.
Moreover, we observe that the image generator $G_i$ outputs a coarse synthesis after the first stage, which yields blurred image details and high pixel-level dissimilarity with the target images. 
Both inspire us to explore a coarse-to-fine generation strategy in order to boost the synthesis performance based on the coarse predictions. 
Cascade models have been used in several other computer vision tasks such as object detection~\cite{chen2014joint} and semantic segmentation~\cite{dai2016instance}, and have shown great effectiveness. 
In this paper, we introduce the cascade strategy to deal with the guided image-to-image translation problems. 
In both stages we have a basic cycled semantic guidance generation sub-network, while in the second stage, we propose two novel multi-scale spatial pooling \& channel selection and multi-channel attention selection modules to better utilize the coarse outputs from the first stage and to produce fine-grained final outputs. We observed significant improvement by using the proposed cascade strategy, illustrated in the experimental part. 

\begin{figure*}[!t] \small
	\centering
	\includegraphics[width=0.85\linewidth]{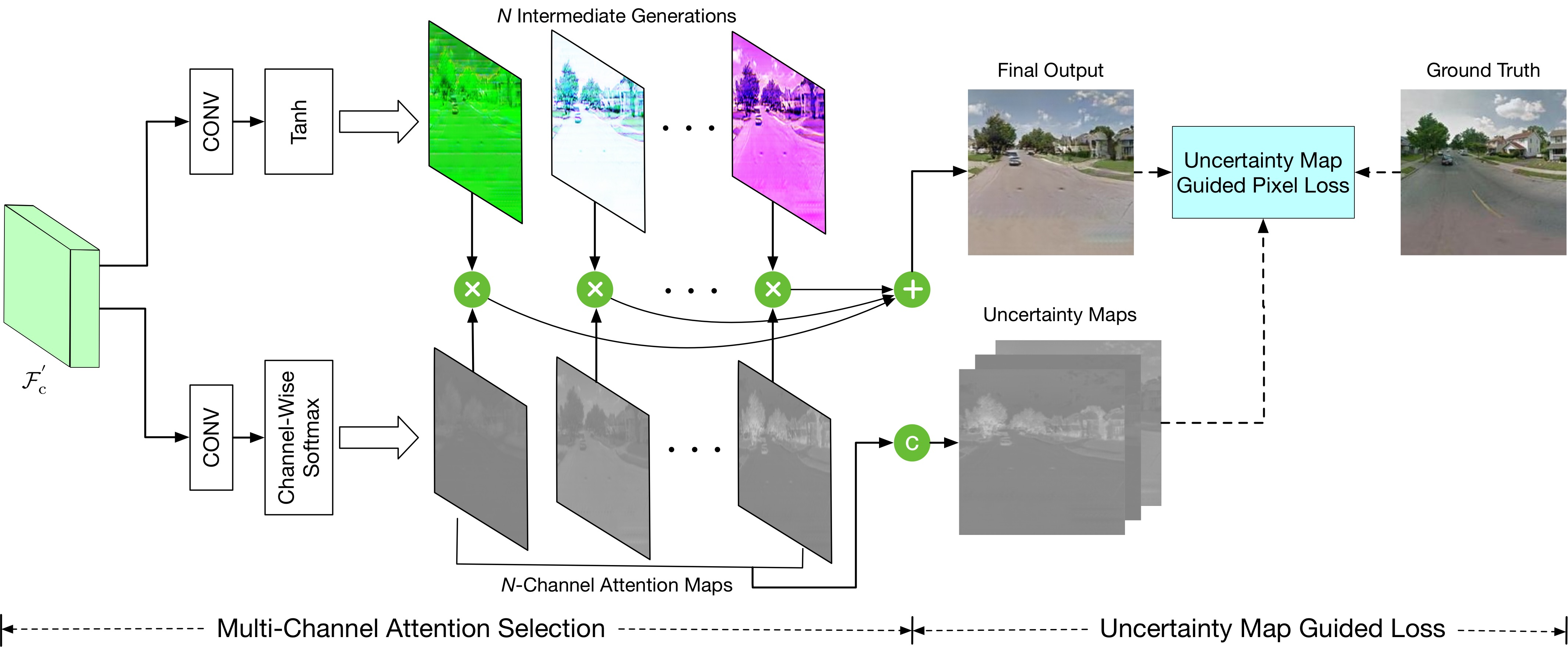}
	\caption{Overview of the proposed multi-channel attention selection module. This module aims to automatically select from a set of intermediate diverse generations in a larger generation space to improve the generation quality. Meanwhile, the module also effectively learns uncertainty maps to guide the pixel loss for robust joint images and guidances optimization. $\oplus$, $\otimes$ and $\textcircled{c}$  denote element-wise addition, element-wise multiplication, and channel-wise concatenation, respectively.}
	\label{fig:selection}
	\vspace{-0.2cm}
\end{figure*}

\subsection{Multi-Scale Spatial Pooling \& Channel Selection}
An overview of the proposed multi-scale spatial pooling \& channel selection module is shown in Fig.~\ref{fig:channel}. 
The module consists of a multi-scale spatial pooling and a multi-scale channel selection components.
In this way, the proposed module can learn multi-scale deep feature interdependencies in both spatial and channel dimensions.

\noindent \textbf{Multi-Scale Spatial Pooling.}
Since there exists a large object/scene deformation between the source and the target domains, a single-scale feature may not be able to capture all the necessary spatial information for a fine-grained generation. Thus, we propose a multi-scale spatial pooling scheme, which uses a set of different kernel sizes and strides to perform a global average pooling on the same input features. By so doing, we obtain multi-scale features with different receptive fields to perceive  different spatial contexts. More specifically, given the coarse inputs and the deep features produced from the stage I, we first concatenate all of them as new features denoted as $\mathcal{F}_{\mathrm{c}} {\in} \mathbb{R}^{C {\times} H {\times} W}$ for the stage II as:

\begin{equation}
\begin{aligned}
\mathcal{F}_{\mathrm{c}} = \mathrm{concat}(I_a, I_g^{'}, F_i, F_s),
\end{aligned}
\end{equation}
where $\mathrm{concat}(\cdot)$ is a function for channel-wise concatenation operation; $F_i$ and $F_s$ are features from the last convolution layers of the generators $G_i$ and $G_s$, respectively. 
$H$ and $W$ are width and height of the features, and $C$ is the number of channels.
We apply a set of $M$ spatial scales $\{s_i\}_{i=1}^M$ in pooling, resulting in pooled features with different spatial resolution.
Different from the pooling scheme used in~\cite{zhao2017pyramid} which directly combines all the features after pooling, we first select each pooled feature via an element-wise multiplication with the input feature. 
Since in our task the input features are from different sources, 
highly correlated features would preserve more useful information for the generation. Let us denote $\mathrm{pl\_up_s}(\cdot)$ as pooling at a scale $s$ followed by an up-sampling operation to rescale the pooled feature at the same resolution, and $\otimes$ as element-wise multiplication, we can formalize the whole process as:
\begin{equation}
\begin{aligned}
\mathcal{F}_{\mathrm{m}} \leftarrow \mathrm{concat}\big(\mathcal{F}_{\mathrm{c}}, \mathcal{F}_{\mathrm{c}} \otimes \mathrm{pl\_up_1}(\mathcal{F}_{\mathrm{c}}),\dots, \mathcal{F}_{\mathrm{c}} \otimes \mathrm{pl\_up}_M(\mathcal{F}_{\mathrm{c}})),
\end{aligned}
\end{equation}
which produces new multi-scale features $\mathcal{F}_{\mathrm{m}} {\in} \mathbb{R}^{4C {\times} H {\times} W}$ (we set $M{=}3$ in our experiments.) for the use in the next multi-scale channel selection module. 
By doing so, the `level' of features can be enriched by combining multiple scale feature maps.

\noindent \textbf{Multi-Scale Channel Selection.} 
Each channel map of $\mathcal{F}_{\mathrm{m}}$ can be now regarded as a scale-specific response, and different scale feature maps should be associated with each other. 
To exploit the interdependencies between each scale features of $\mathcal{F}_{\mathrm{m}}$, we propose a multi-scale channel selection module to explicitly model interdependencies between channels of multi-scale feature $\mathcal{F}_{\mathrm{m}}$.
The structure of multi-scale channel selection module is illustrated in Fig.~\ref{fig:channel}.

The channel attention map $\mathcal{A}$ can be obtained from the multi-scale feature $\mathcal{F}_{\mathrm{m}}$.
More specific, $\mathcal{F}_{\mathrm{m}}$ is first reshaped to $\mathbb{R}^{4C {\times} HW}$, and then a matrix multiplication is preformed between $\mathcal{F}_{\mathrm{m}}$ and the transpose of $\mathcal{F}_{\mathrm{m}}$.
Next, we employ a Softmax activation function to obtain the channel attention map $\mathcal{A} {\in} \mathbb{R}^{4C {\times} 4C}$.
Each pixel $\mathcal{A}_{ji}$ in $\mathcal{A}$  measures the $i^{th}$ channel's impact on the $j^{th}$ channel.
In this way, the correlation can be built between features from different scales.
Moreover, to reshape back to $\mathbb{R}^{4C {\times} H {\times} W}$, we perform a matrix multiplication between $\mathcal{A}$ and the transpose of $\mathcal{F}_{\mathrm{m}}$.
Then, the result is multiplied by a parameter $\alpha$ and added to the original feature $\mathcal{F}_{\mathrm{m}}$ to obtain the channel-wise enhanced feature $\mathcal{F}_{\mathrm{m}}^{'} {\in} \mathbb{R}^{4C {\times} H {\times} W}$,
\begin{equation}
\begin{aligned}
\mathcal{F}_{\mathrm{m}}^{'} = \alpha \sum_{i=1}^{4C} (\mathcal{A}_{ji} {\mathcal{F}_{\mathrm{m}}}_{i}) + {\mathcal{F}_{\mathrm{m}}}_{j}.
\end{aligned}
\end{equation}
By doing so, each channel in the final feature $\mathcal{F}_{\mathrm{m}}^{'}$ is a weighted sum of all channels and it models the long-range dependencies between multi-scale feature maps.
Finally, the enhanced feature $\mathcal{F}_{\mathrm{m}}^{'}$ is fed into a convolutional layer to obtain $\mathcal{F}_{\mathrm{c}}^{'} {\in} \mathbb{R}^{C {\times} H {\times} W}$, which has the same size as the original one $\mathcal{F}_{\mathrm{c}}$.
This design ensures that the proposed multi-scale spatial pooling \& channel selection module can be plugged into existing computer vision architectures.

\subsection{Multi-Channel Attention Selection}
In previous image-to-image translation works, the image was generated only in a three-channel RGB space. We argue that this is not enough for the complex translation problem we are dealing with, and thus we explore using a larger generation space to have a richer synthesis via constructing multiple intermediate generations.
Accordingly, we design a multi-channel attention mechanism to automatically perform spatial and temporal selection from the generations to synthesize a fine-grained final output.

Given the enhanced multi-scale feature volume $\mathcal{F}_{\mathrm{c}}^{'}{\in} \mathbb{R}^{C{\times} H {\times} W}$, where $H$ and $W$ are width and height of the features, and $C$ is the number of channels, we consider two directions as shown in Fig.~\ref{fig:selection}. 
One is for the generation of multiple intermediate image synthesis and the other is for the generation of multi-channel attention maps. To produce $N$ different intermediate generations $I_G{=}\{I_G^i\}_{i=1}^N$, a convolution operation is performed with $N$ convolutional filters $\{W_G^i,b_G^i\}_{i=1}^N$ followed by a $\tanh(\cdot)$ non-linear activation operation. For the generation of corresponding $N$ attention maps, the other group of filters $\{W_A^i,b_A^i\}_{i=1}^N$ is applied. Then the intermediate generations and the attention maps are calculated as follows:
\begin{equation} 
\begin{aligned}
I_G^i = \tanh(\mathcal{F}_{\mathrm{c}}^{'}W_G^i + b_G^i), \quad \quad \text{for} \ \ i = 1, \dots, N \\
I_A^i = \mathrm{Softmax}(\mathcal{F}_{\mathrm{c}}^{'}W_A^i + b_A^i), \quad \quad \text{for} \ \ i = 1, \dots, N 
\end{aligned}
\label{eqn:attention}
%\vspace{-0.1cm}
\end{equation}
where $\mathrm{Softmax}(\cdot)$ is a channel-wise softmax function used for the normalization. Finally, the learned attention maps are utilized to perform channel-wise selection from each intermediate generation as follows:
\begin{equation}
\begin{aligned}
I_g^{''} = (I_A^1 \otimes I_G^1) \oplus \cdots \oplus (I_A^N \otimes I_G^N) 
\end{aligned}
\end{equation}
where $I_g^{''}$ represents the final synthesized generation selected from the multiple diverse results, and  $\oplus$ denotes the element-wise addition.
We also generate a final semantic guidance in the second stage as in the first stage, i.e., $
S_g^{''}  {=} G_s(I_g^{''})$. Due to the same purpose of the two semantic guidance generators, we use a single $G_s$ twice by sharing the parameters in both stages to reduce the network capacity.

\noindent \textbf{Uncertainty-Guided Pixel Loss.}
As we discussed in the introduction, the semantic guidance obtained from the pretrained model is not accurate for all the pixels, leading to a wrong guidance during training. To tackle this issue, we propose to learn uncertainty maps to control the optimization loss as shown in Fig.~\ref{fig:selection}. The uncertainty learning has been investigated in~\cite{kendall2017multi} for multi-task learning, and here we introduce it for solving the noisy semantic guidance problem.
Assume that we have $K$ different loss maps which need a guidance. The multiple generated attention maps are first concatenated and passed to a convolution layer with $K$ filters $\{W_u^i\}_{i=1}^K$ to produce a set of $K$ uncertainty maps. The reason for using the attention maps to generate uncertainty maps is that the attention maps directly affect the final generation leading to a close connection with the loss. Let $\mathcal{L}_p^i$ denote a pixel-level loss map and $U_i$ denote the $i$-th uncertainty map, we have:
\begin{equation} \small
\begin{aligned}
&U_i = \sigma\big(W_u^i(\mathrm{concat}(I_A^1,\dots,I_A^N) + b_u^i\big) \\
&\mathcal{L}_{p}^i \leftarrow \frac{\mathcal{L}_{p}^i }{U_i}+ \log U_i, \quad \text{for} \ \  i=1,\dots,K
\end{aligned}
\end{equation}
where $\sigma(\cdot)$ is a Sigmoid function for pixel-level normalization. The uncertainty map is automatically learned and acts as a weighting scheme to control the optimization loss.

\noindent \textbf{Parameter-Sharing Discriminator.}
We extend the vanilla discriminator in~\cite{isola2017image} to a parameter-sharing structure. 
In the first stage, this structure takes the real image $I_a$ and the generated image $I_g^{'}$ or the ground-truth image $I_g$ as input. The discriminator $D$ learns to tell whether a pair of images from different domains is associated with each other or not. In the second stage, it accepts the real image $I_a$ and the generated image $I_g^{''}$ or the real image $I_g$ as inputs. This pairwise input encourages $D$ to discriminate the diversity of image structure and to capture the local-aware information. 

\subsection{Overall Optimization Objective}
\noindent \textbf{Adversarial Loss.}
In the first stage, the adversarial loss of $D$ for distinguishing synthesized image pairs $[I_a, I_g^{'}]$ from real image pairs $[I_a, I_g]$ is formulated as follows:
\begin{equation}
\begin{aligned}
\mathcal{L}_{cGAN}(I_a, I_g^{'}) = 
& \mathbb{E}_{I_a, I_g} \left[ \log D(I_a, I_g) \right] +  \\
& \mathbb{E}_{I_a, I_g^{'}} \left[\log (1 - D(I_a, I_g^{'})) \right].
\end{aligned}
\label{eqn:adv_1}
\end{equation}
In the second stage, the adversarial loss of $D$ for distinguishing synthesized image pairs $[I_a, I_g^{''}]$ from real image pairs $[I_a, I_g]$ is formulated as follows:
\begin{equation} 
\begin{aligned}
\mathcal{L}_{cGAN}(I_a, I_g^{''}) {=} 
& \mathbb{E}_{I_a, I_g} \left[ \log D(I_a, I_g) \right] +  \\
& \mathbb{E}_{I_a, I_g^{''}} \left[\log (1 - D(I_a, I_g^{''})) \right].
\end{aligned}
\label{eqn:adv_2}
\end{equation}
Both losses aim to preserve the local structure information and produce visually pleasing synthesized images.
Thus, the adversarial loss of the proposed SelectionGAN is the sum of Eq.~\eqref{eqn:adv_1} and  \eqref{eqn:adv_2},
\begin{equation}
\begin{aligned}
\mathcal{L}_{cGAN} =  \mathcal{L}_{cGAN}(I_a, I_g^{'}) +  \lambda \mathcal{L}_{cGAN}(I_a, I_g^{''}).
\end{aligned}
\label{eq:adv}
%\vspace{-0.1cm}
\end{equation}

\noindent \textbf{Overall Loss.}
The total optimization loss is a weighted sum of the above losses.
Generators $G_i$, $G_s$, multi-scale spatial pooling \& channel selection module $G_m$, multi-channel attention selection network $G_a$, and discriminator $D$ are trained in an end-to-end fashion optimizing the following min-max function:
\begin{equation}
\begin{aligned}
\min_{\{G_i, G_s, G_m, G_a\}} \max_{\{D\}} \mathcal{L} = & \sum_{i=1}^4 \lambda_i \mathcal{L}_{p}^i + \mathcal{L}_{cGAN} + \lambda_{tv}\mathcal{L}_{tv}.
\end{aligned}
\label{eqn:all}
\end{equation} 
where $\mathcal{L}_p^i$ uses the L1 reconstruction to separately calculate the pixel loss between the generated four images/guidances (i.e., $I_g^{'}$, $S_g^{'}$, $I_g^{''}$, and $S_g^{''}$) and the corresponding real images/guidances. 
$\mathcal{L}_{tv}$ is the total variation regularization~\cite{johnson2016perceptual} on the final synthesized image $I_g^{''}$.
$\lambda_i$ and $\lambda_{tv}$ are the trade-off parameters to control the relative importance of different objectives. 
The training is performed by solving the min-max optimization problem.

\begin{table*}[!t] \small
\centering
\caption{Ablations study of the proposed SelectionGAN.}
\resizebox{0.9\linewidth}{!}{
	\begin{tabular}{ccccccccc} \toprule
		\multirow{2}{*}{Baseline}  & \multirow{2}{*}{Setups of SelectionGAN} & \multirow{2}{*}{SSIM $\uparrow$}   & \multirow{2}{*}{PSNR $\uparrow$} & \multirow{2}{*}{SD $\uparrow$} & \multirow{2}{*}{FID $\downarrow$} & \multicolumn{3}{c}{Inception Score $\uparrow$} \\ \cmidrule(lr){7-9} 
		& & & & & & All & Top-1 & Top-5 \\ \hline
		A	          & $I_a \stackrel{G_i} \rightarrow I_g^{'}$ & 0.4555  & 19.6574 & 18.8870  & 91.47 & 3.2359 & 2.1903 & 3.3110                     \\
		B & $ S_g \stackrel{G_i} \rightarrow I_g^{'}$  & 0.5223 & 22.4961 & 19.2648 & 87.51 & 3.4849 & 2.2544 & 3.4217 \\
		C	          & $[I_a, S_g] \stackrel{G_i} \rightarrow I_g^{'}$ & 0.5374  & 22.8345 & 19.2075 & 84.10 & 3.4478 & 2.2616 & 3.4668             \\
		D             & $[I_a, S_g] \stackrel{G_i} \rightarrow I_g^{'} \stackrel{G_s} \rightarrow S_g^{'}$ & 0.5438 & 22.9773 & 19.4568 & 82.81 & 3.1655 & 2.2561 & 3.2401\\
		E             & D + Uncertainty-Guided Pixel Loss                & 0.5522 & 23.0317 & 19.5127 & 79.84 & 3.2741 & 2.2687 & 3.3063 \\
		F             & E + Multi-Channel Attention Selection            & 0.5989 & 23.7562 & 20.0000 & 75.57 & 3.3365 & \textbf{2.2749} & 3.4664 \\
		G             & F + Total Variation Regularization               & 0.6047  & 23.7956 & 20.0830 & 74.11 & 3.3172 & 2.1397 & 3.3509 \\
		H             & G + Multi-Scale Spatial Pooling                  & \textbf{0.6167} & \textbf{23.9310} & \textbf{20.1214} & \textbf{72.23} & \textbf{3.4978} & 2.1880 & \textbf{3.4786} \\
		\bottomrule		
	\end{tabular}}
	\vspace{-0.2cm}
	\label{tab:ablation}
\end{table*}

\subsection{Implementation Details}
\noindent \textbf{Network Architecture.}
For a fair comparison, we employ U-Net~\cite{isola2017image} as our generator architectures $G_i$ and $G_s$.
U-Net is a network with skip connections between a down-sampling encoder and an up-sampling decoder. 
Such architecture comprehensively retains contextual and textural information, which is crucial for removing artifacts and padding textures. 
Since our focus is on the image generation task, $G_i$ is more important than $G_s$.
Thus we use a deeper network for $G_i$ and a shallow network for $G_s$. 
Specifically, the filters in first convolutional layer of $G_i$ and $G_s$ are 64 and 4, respectively.
For the network $G_a$, the kernel size of convolutions for generating the intermediate images and attention maps are $3{\times}3$ and $1{\times }1$, respectively.
We adopt PatchGAN~\cite{isola2017image} for the discriminator $D$.

\noindent \textbf{Training Details.}
We mainly focus on four guided image-to-image translation tasks in this paper.
For cross-view image translation, we follow~\cite{regmi2018cross} and use RefineNet \cite{lin2017refinenet} and \cite{zhou2017scene} to generate segmentation maps on Dayton, SVA, and Ego2Top datasets as training data, respectively.
For facial expression generation, we follow~\cite{tang2019cycle} and use OpenFace \cite{amos2016openface} to extract facial landmarks on Radboud Faces dataset as training data.
For both hand gesture generation and human pose generation tasks, we follow \cite{tang2018gesturegan,ma2017pose} and employ OpenPose \cite{cao2017realtime} as pose joints detector and filter out images where no human hand and body are detected in the associated datasets. 

We follow the optimization method in~\cite{goodfellow2014generative} to optimize the proposed SelectionGAN, i.e., one gradient descent step on discriminator and generators alternately.
We first train $G_i$, $G_s$, $G_m$, $G_a$ with $D$ fixed, and then train $D$ with $G_i$, $G_s$, $G_m$, $G_a$  fixed.
The proposed SelectionGAN is trained and optimized in an end-to-end fashion.
We employ  Adam \cite{kingma2014adam} with momentum terms  $\beta_1{=}0.5$ and $\beta_2{=}0.999$ as our solver.
In our experiments, we set $\lambda_{tv}$=$1e{-}6$, $\lambda_1{=}100$, $\lambda_2{=}1$, $\lambda_3{=}200$ and $\lambda_4{=}2$ in Eq.~\eqref{eqn:all}, and $\lambda{=}4$ in Eq.~\eqref{eq:adv}.
The number of attention channels $N$ in Eq.~\eqref{eqn:attention} is set to 10.
\section{Experiments}
\label{sec:experiment}
We conduct extensive experiments on a variety of guided image-to-image translation tasks such as segmentation map guided cross-view image translation, 
facial landmark guided expression-to-expression translation, hand skeleton guided gesture-to-gesture translation, and pose skeleton guided person image generation.
Moreover, to explore the generality of the proposed SelectionGAN on other generation tasks, we conduct experiments on the challenging semantic image synthesis task.

\subsection{Results on Cross-View Image Translation}

\noindent \textbf{Datasets.}
We follow~\cite{regmi2018cross,regmi2019cross,tang2019multi} and perform experiments on four public cross-view image translation datasets: 
1) The Dayton dataset~\cite{vo2016localizing} contains 76,048 images and the train/test split is 55,000/21,048. 
The images in the original dataset have $354{\times}354$ resolution. 
We resize them to $256{\times}256$.
2) The CVUSA dataset \cite{workman2015wide} consists of 35,532/8,884 image pairs in train/test split. 
Following~\cite{zhai2017predicting,regmi2018cross}, the aerial images are center-cropped to $224{\times}224$ and resized to $256{\times}256$. 
For the ground level images and corresponding segmentation maps, we take the first quarter of both and resize them to $256{\times}256$.
3) The Surround Vehicle Awareness (SVA) dataset~\cite{palazzi2017learning} is a synthetic dataset collected from Grand Theft Auto V (GTAV) video game. 
Following~\cite{regmi2019cross}, we select every tenth frame to remove redundancy in this dataset since the consecutive frames in each set are very similar to each other. 
Thus, we collect 46,030/22,254 image pairs for training and testing, respectively. 
4) The Ego2Top dataset~\cite{ardeshir2016ego2top} is more challenging and contains different indoor and outdoor conditions.
Each case contains one top-view video and several egocentric videos captured by the people visible in the top-view camera. This dataset has more than 230,000 frames. For training data, we follow \cite{tang2019multi} and randomly select 386,357 pairs and each pair is composed of two images of the same scene but different viewpoints. We randomly select 25,600 pairs for evaluation.

\noindent \textbf{Parameter Settings.}
For a fair comparison, we adopt the same training setup as in~\cite{isola2017image,regmi2018cross}. All images are scaled to $256{\times}256$, and we enabled image flipping and random crops for data augmentation.
Similar to~\cite{regmi2018cross}, the experiments for Dayton are trained for 35 epochs with a batch size of 4.
For CVUSA, we follow the same setup as in~\cite{zhai2017predicting,regmi2018cross}, and train our network for 30 epochs with batch size of 4.
For SVA, all models are trained with  20 epoch using batch size~4.

\noindent \textbf{Evaluation Metrics.}
Similar to~\cite{regmi2018cross,tang2019multi}, we employ Inception Score~\cite{salimans2016improved}, top-k prediction accuracy, KL score, and Fr\'echet Inception Distance (FID)~\cite{heusel2017gans} for the quantitative analysis.
These metrics evaluate the generated images from a high-level feature space. We also employ pixel-level similarity metrics to evaluate our method, i.e., Structural-Similarity (SSIM) \cite{wang2004image}, Peak Signal-to-Noise Ratio (PSNR), and Sharpness Difference (SD).

\noindent\textbf{Baseline Models.} 
We first conduct an ablation study on Dayton to evaluate the components of the proposed SelectionGAN. 
To reduce the training time, we randomly select 1/3 samples from the whole 55,000/21,048 samples, i.e., around 18,334 samples for training and 7,017 samples for testing. 
The proposed SelectionGAN considers eight baselines (A, B, C, D, E, F, G, H) as shown in Table~\ref{tab:ablation}. 
Baseline A uses a Pix2pix structure~\cite{isola2017image} and generates $I_g^{'}$ using a single image $I_a$. 
Baseline B uses the same Pix2pix model and generates $I_g^{'}$ using the corresponding semantic guidance $S_g$.
Baseline C also uses the Pix2pix structure, and inputs the combination of a conditional image $I_a$ and the semantic guidance $S_g$ to the generator $G_i$.
Baseline D uses the proposed cycled semantic guidance generation upon Baseline C.
Baseline E represents the pixel loss guided by the learned uncertainty maps.
Baseline F employs the proposed multi-channel attention selection module to generate multiple intermediate generations, and to make the neural network attentively select which part is more important for generating the target image. 
Baseline G adds the total variation regularization on the final result $I_g^{''}$.
Baseline H employs the proposed multi-scale spatial pooling module to refine the features $\mathcal{F}_{\mathrm{c}}$ from stage I. All the baseline models are trained and tested on the same data using the configuration.

\begin{figure}[!t]\small
	\centering
	\includegraphics[width=.9\linewidth]{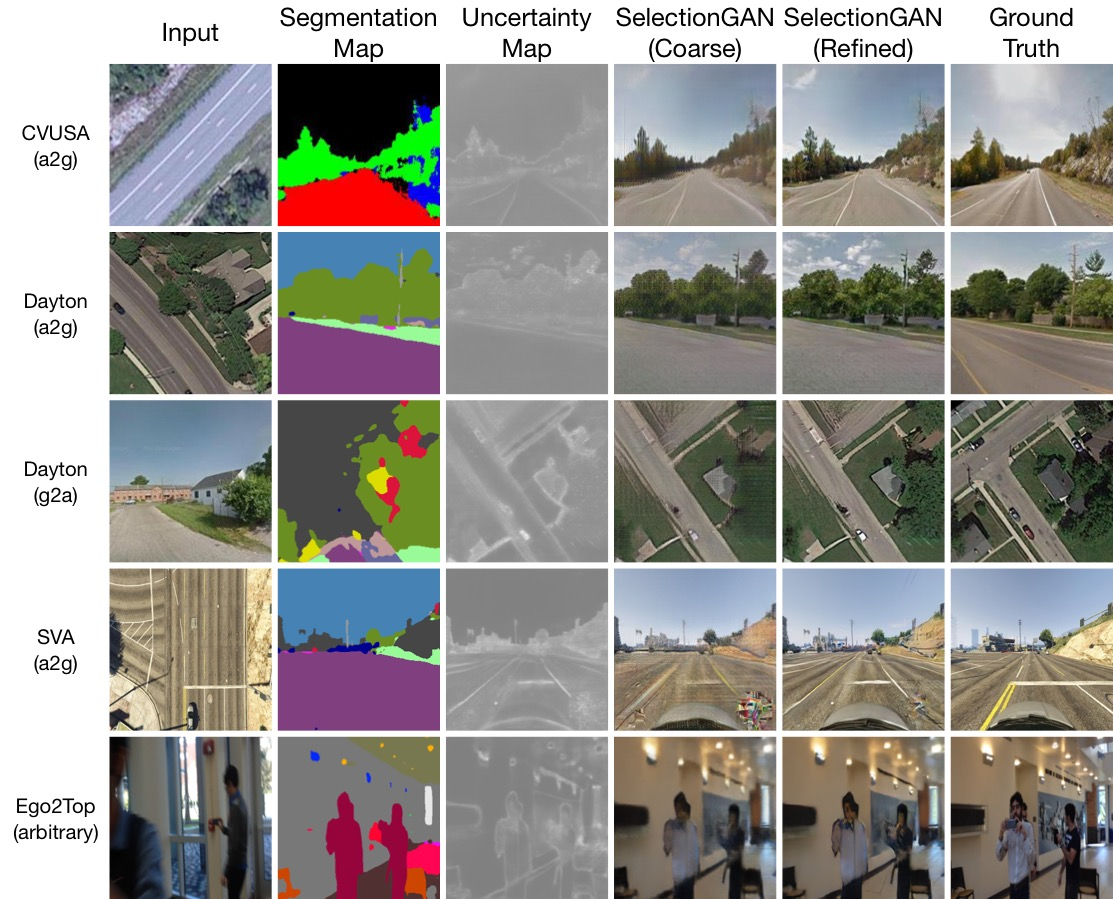}
	\caption{Results of cross-view image translation generated by the proposed SelectionGAN on different datasets.}
	\label{fig:cross_view_coarse_fine}
	\vspace{-0.2cm}
\end{figure}

\begin{table}[!t] \small
\centering
\caption{Quantitative results of coarse-to-fine generation.} 
	\begin{tabular}{cccccc} \toprule
		\#   & Stage I    & Stage II & SSIM    & PSNR    & SD           \\	\midrule
		F          &  $\surd$   &          & 0.5551  & 23.1919  & 19.6311  \\
		F          &            & $\surd$  & \textbf{0.5989} & \textbf{23.7562} & \textbf{20.0000}  \\ 
		G          &  $\surd$   &          &  0.5680   & 23.2574 & 19.7371  \\
		G          &            & $\surd$  & \textbf{0.6047}  & \textbf{23.7956} & \textbf{20.0830} \\
		H          &  $\surd$   &          &  0.5567 & 23.1545 & 19.6034    \\
		H          &            & $\surd$  &  \textbf{0.6167} & \textbf{23.9310} & \textbf{20.1214}    \\ 
		\bottomrule		
	\end{tabular}
	\label{tab:cos}
	\vspace{-0.2cm}
\end{table}

\begin{figure}[!t] \small
	\centering
	\includegraphics[width=1\linewidth]{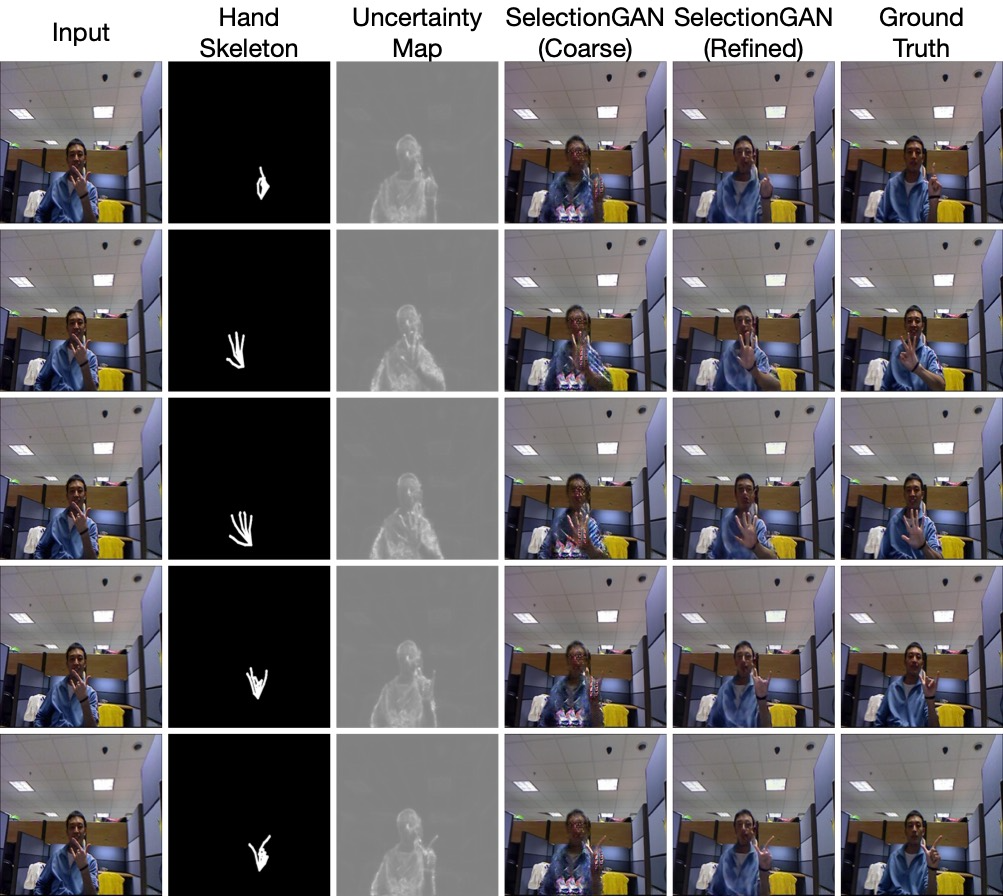}
	\caption{Results of controllable hand gesture translation.}
	\label{fig:gesture_results}
	\vspace{-0.4cm}
\end{figure}

\noindent \textbf{Ablation Analysis.}
The results of the ablation study are shown in Table~\ref{tab:ablation}. 
Note that Baseline B is better than A since $S_g$ contains more structural information than $I_a$. 
When comparing Baselines A and C, the semantic-guided generation improves SSIM, PSNR and SD by 8.19, 3.1771 and 0.3205, respectively, confirming the importance of the conditional semantic guidance information.
By using the proposed cycled semantic guidance generation, Baseline D further improves over C, meaning that the proposed semantic guidance cycle structure indeed utilizes the semantic guidance information in a more effective way, confirming our design motivation.
Baseline E outperforms D showing the importance of using the uncertainty maps to guide the pixel loss map which contains an inaccurate reconstruction loss due to the wrong semantic guidance produced from the pre-trained models.
Baseline F significantly outperforms E with around 4.67 points gain on the SSIM metric, clearly demonstrating the effectiveness of the proposed multi-channel attention selection scheme.
We can also observe from Table~\ref{tab:ablation} that, by adding the proposed multi-scale spatial pool scheme and the TV regularization, the overall performance is further boosted.
Finally, we demonstrate the advantage of the proposed two-stage strategy over the one-stage method. 
The results are shown in Fig.~\ref{fig:cross_view_coarse_fine}, \ref{fig:gesture_results}, and Table~\ref{tab:cos}. 
It is obvious that the coarse-to-fine generation model is able to generate sharper results and contains more details than the one-stage model, which further confirms our motivations.

\begin{figure}[!t]\small
	\centering
	\includegraphics[width=0.85\linewidth]{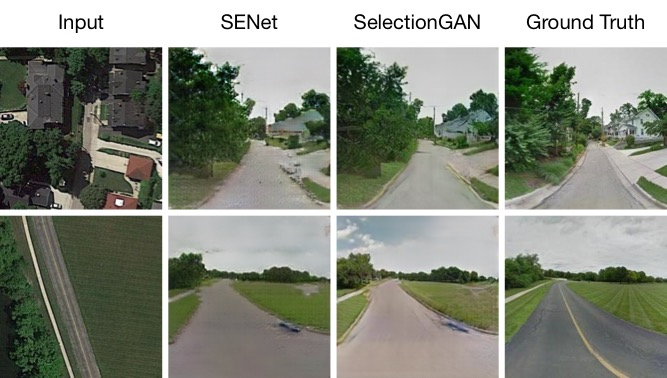}
	\caption{Comparison results of SENet and the proposed SelectionGAN on Dayton.}
	\label{fig:senet}
	\vspace{-0.2cm}
\end{figure}

\noindent \textbf{Comparisons with SENet \cite{hu2018squeeze}.}
The proposed multi-scale spatial pooling shares a similar intuition with SENet \cite{hu2018squeeze} which amplifies the channels via attention based on pooling. 
Unlike SENet that employs positive attention via the Sigmoid function, the proposed multi-scale spatial pooling selects each pooled feature via an element-wise multiplication with the original feature.
Since in our task the input features are from different sources, highly correlated features would preserve more useful information for the generation.
We also conduct experiments to compare the proposed method with SENet on Dayton. Specifically, we use the SE layer proposed in \cite{hu2018squeeze} to replace our multi-scale spatial pooling module, obtaining the following results in terms of SSIM, PSNR, and SD: 0.5912, 23.3857, and 19.8061, respectively. We can see that our method (see the Baseline H in Table~\ref{tab:ablation}) still significantly outperforms \cite{hu2018squeeze}.
Moreover, we provide the visualization results in Fig.~\ref{fig:senet} (note that our method achieves better results than SENet).

\begin{table}[!t] \small
\centering
\caption{Influence of the number of attention channels $N$.} 
	\begin{tabular}{cccc} \toprule
		$N$   &  SSIM    & PSNR    & SD \\	\midrule	
		0     & 0.5438 & 22.9773 & 19.4568  \\
		1     & 0.5522 & 23.0317 & 19.5127 \\ 
		5     &  0.5901  & 23.8068 & \textbf{20.0033} \\
		10    & \textbf{0.5986}  & 23.7336 & 19.9993 \\
		32	  & 0.5950  & \textbf{23.8265} & 19.9086 \\ 
		\bottomrule		
	\end{tabular}
	\vspace{-0.2cm}
	\label{tab:attention_n}
\end{table}

\begin{figure*}[!t] \small
	\centering
	\includegraphics[width=0.85\linewidth]{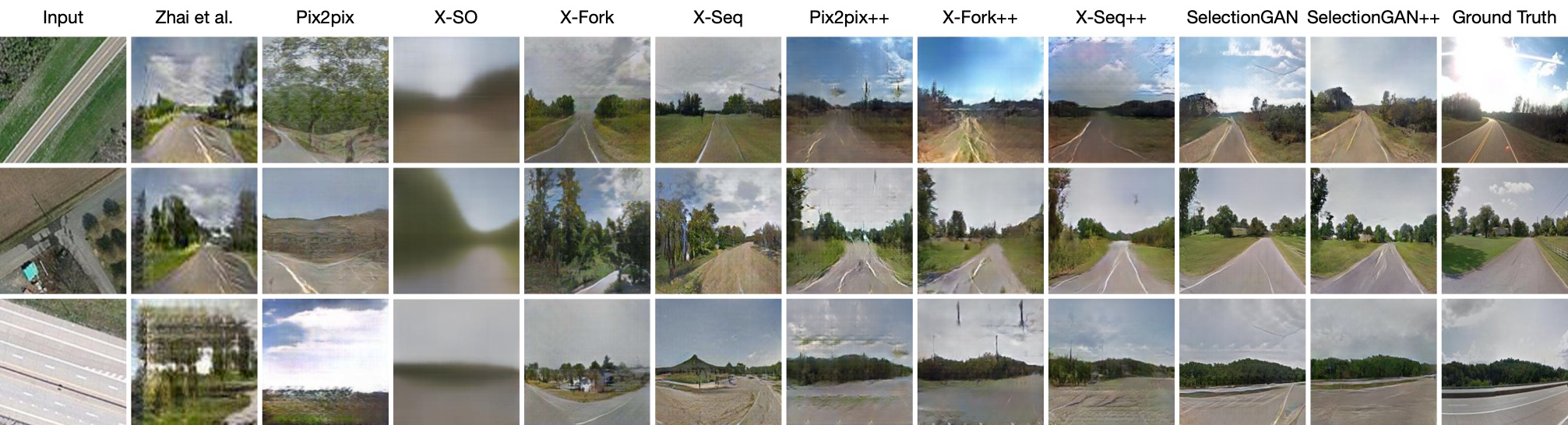}
	\caption{Results of cross-view image translation on CVUSA.}
	\label{fig:cvusa_dayton_ego2top}
	\vspace{-0.2cm}
\end{figure*}

\begin{figure*}[!t] \small
	\centering
	\includegraphics[width=0.85\linewidth]{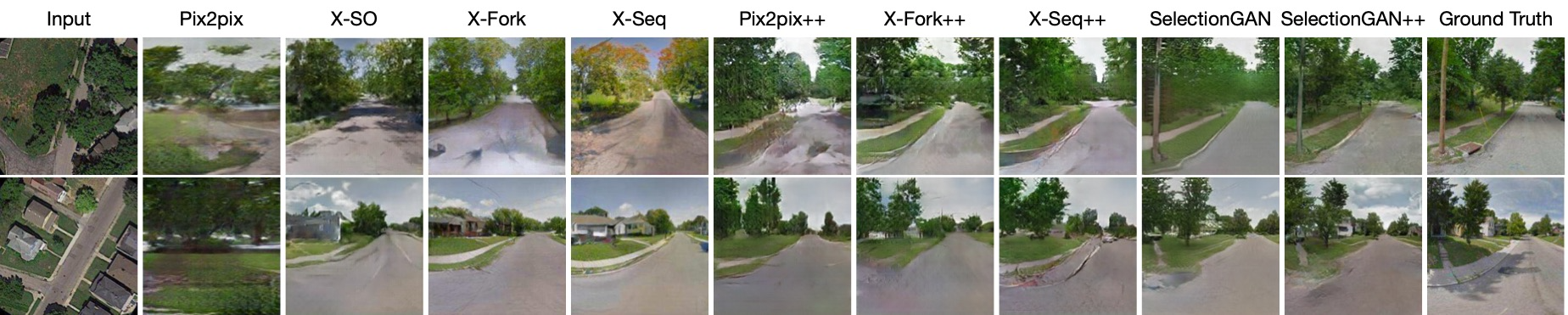}
	\caption{Results of cross-view image translation on Dayton in a2g direction.}
	\label{fig:dayton}
	\vspace{-0.2cm}
\end{figure*}

\noindent \textbf{Influence of the Number of Attention Channels.}
We investigate the influence of the number of attention channel $N$ in Eq.~\eqref{eqn:attention}.
The results are shown in Table \ref{tab:attention_n}.
We observe that the performance tends to be stable after $N{=}10$.
Thus, taking both performance and training speed into consideration, we set $N{=}10$ in all our experiments.

\begin{table*}[!tbp] \small
	\centering
	\caption{Quantitative results of cross-view image translation on CVUSA. ($\ast$) Inception Score for real (ground truth) data is 4.8741, 3.2959 and 4.9943 for all, top-1 and top-5 setups, respectively.}
		\resizebox{0.9\linewidth}{!}{% 
		\begin{tabular}{lcccccccccccc} \toprule
			\multirow{2}{*}{Method} & \multicolumn{4}{c}{Accuracy (\%) $\uparrow$}& \multicolumn{3}{c}{Inception Score$^\ast$ $\uparrow$} & \multirow{2}{*}{SSIM $\uparrow$} & \multirow{2}{*}{PSNR $\uparrow$} & \multirow{2}{*}{SD $\uparrow$} & \multirow{2}{*}{KL $\downarrow$} \\ \cmidrule(lr){2-5} \cmidrule(lr){6-8} 
			& \multicolumn{2}{c}{Top-1} & \multicolumn{2}{c}{Top-5} & All & Top-1 & Top-5  \\ \midrule
			Zhai et al. \cite{zhai2017predicting} &13.97 &14.03 &42.09 &52.29 & 1.8434 &1.5171 &1.8666  & 0.4147 &17.4886  &16.6184  & 27.43 $\pm$ 1.63  \\
			Pix2pix~\cite{isola2017image}        &7.33  &9.25  &25.81 &32.67 & 3.2771 &2.2219 &3.4312  & 0.3923 &17.6578  &18.5239  & 59.81 $\pm$ 2.12   \\ 
			X-SO~\cite{regmi2019cross}           &0.29  &0.21  & 6.14 & 9.08 & 1.7575 &1.4145 &1.7791  & 0.3451 & 17.6201 & 16.9919 & 414.25 $\pm$ 2.37 \\
			X-Fork~\cite{regmi2018cross}         &20.58 &31.24 &50.51 &63.66 & 3.4432 &2.5447 &3.5567  & 0.4356 &19.0509  &18.6706  & 11.71 $\pm$ 1.55 \\
			X-Seq~\cite{regmi2018cross}          &15.98 &24.14 &42.91 &54.41 & 3.8151 &2.6738 & 4.0077  & 0.4231  &18.8067  &18.4378  &15.52 $\pm$ 1.73  \\ 
			Pix2pix++~\cite{isola2017image}      &26.45 &41.87 &57.26 &72.87 & 3.2592 &2.4175 &3.5078 & 0.4617 & 21.5739 & 18.9044 & 9.47 $\pm$ 1.69 \\
			X-Fork++~\cite{regmi2018cross}       &31.03 &49.65 &64.47 &81.16 & 3.3758 &2.5375 &3.5711 & 0.4769 & 21.6504 & 18.9856 &7.18 $\pm$ 1.56\\
			X-Seq++~\cite{regmi2018cross}        &34.69 &54.61 &67.12 &83.46 & 3.3919 &2.5474 &3.4858 & 0.4740 & 21.6733 & 18.9907 & 5.19 $\pm$ 1.31\\
			SelectionGAN                         & 41.52 & 65.51 & 74.32 & 89.66 & 3.8074 & 2.7181 &3.9197 & 0.5323  & \textbf{23.1466}  & 19.6100  & 2.96 $\pm$ 0.97 \\
			SelectionGAN++ & \textbf{43.27} & \textbf{68.36} & \textbf{77.15} & \textbf{91.74} & \textbf{3.8296} & \textbf{2.8977} & \textbf{4.0238} & \textbf{0.5355} & 22.8532 & \textbf{19.7672} & \textbf{2.76 $\pm$ 0.96}\\           	
			\bottomrule		
		\end{tabular}}
		\label{tab:cvusa}
		\vspace{-0.2cm}
	\end{table*}
		
\begin{table*}[!ht] \small
	\centering
	\caption{Quantitative results of cross-view image translation on Dayton in a2g direction. ($\ast$) Inception Score for real (ground truth) data is 3.8319, 2.5753 and 3.9222 for all, top-1 and top-5 setups, respectively.}
	\resizebox{0.9\linewidth}{!}{% 
		\begin{tabular}{lccccccccccc} \toprule
			\multirow{2}{*}{Method} & \multicolumn{4}{c}{Accuracy (\%) $\uparrow$} & \multicolumn{3}{c}{Inception Score$^\ast$ $\uparrow$} & \multirow{2}{*}{SSIM $\uparrow$} & \multirow{2}{*}{PSNR $\uparrow$} & \multirow{2}{*}{SD $\uparrow$} & \multirow{2}{*}{KL $\downarrow$}  \\ \cmidrule(lr){2-5} \cmidrule(lr){6-8} 
			& \multicolumn{2}{c}{Top-1} & \multicolumn{2}{c}{Top-5} & All & Top-1 & Top-5 \\ \midrule
			Pix2pix \cite{isola2017image}     &6.80  &9.15 &23.55&27.00& 2.8515&1.9342&2.9083 & 0.4180 &17.6291&19.2821& 38.26 $\pm$ 1.88 \\
			X-SO \cite{regmi2019cross}        &27.56 & 41.15 & 57.96 & 73.20 & 2.9459 & 2.0963 & 2.9980 & 0.4772 & 19.6203 & 19.2939 & 7.20 $\pm$ 1.37 \\
			X-Fork \cite{regmi2018cross}      &30.00 &48.68&61.57&78.84& 3.0720&2.2402&3.0932 &0.4963&19.8928&19.4533  &6.00 $\pm$ 1.28 \\
			X-Seq \cite{regmi2018cross}       &30.16 &49.85&62.59&80.70& 2.7384&2.1304&2.7674 &0.5031 &20.2803 &19.5258 & 5.93 $\pm$ 1.32 \\
			Pix2pix++~\cite{isola2017image}   &32.06 &54.70&63.19&81.01& 3.1709 &2.1200&3.2001&0.4871&21.6675&18.8504& 5.49 $\pm$ 1.25\\
			X-Fork++~\cite{regmi2018cross}    &34.67 & 59.14 &66.37&84.70&3.0737&2.1508&3.0893&0.4982&21.7260&18.9402& 4.59 $\pm$ 1.16 \\
			X-Seq++~\cite{regmi2018cross}     &31.58 & 51.67 &65.21 & 82.48 &3.1703&2.2185& 3.2444 &0.4912&21.7659&18.9265& 4.94 $\pm$ 1.18 \\
			SelectionGAN                      & 42.11 & 68.12 & 77.74 & 92.89 & 3.0613 & 2.2707 & 3.1336 & \textbf{0.5938} & \textbf{23.8874} & \textbf{20.0174} & 2.74 $\pm$ 0.86 \\ 
			SelectionGAN++ & \textbf{47.01} & \textbf{73.54} & \textbf{80.19} & \textbf{94.97} & \textbf{3.2315} & \textbf{2.3367} & \textbf{3.3245} & 0.5786 & 23.5385 & 19.8729 &\textbf{ 2.45 $\pm$ 0.83}\\ \bottomrule	
		\end{tabular}}
		\label{tab:dayton}
		\vspace{-0.4cm}
\end{table*}

\begin{figure*}[!t] \small
	\centering
	\includegraphics[width=0.85\linewidth]{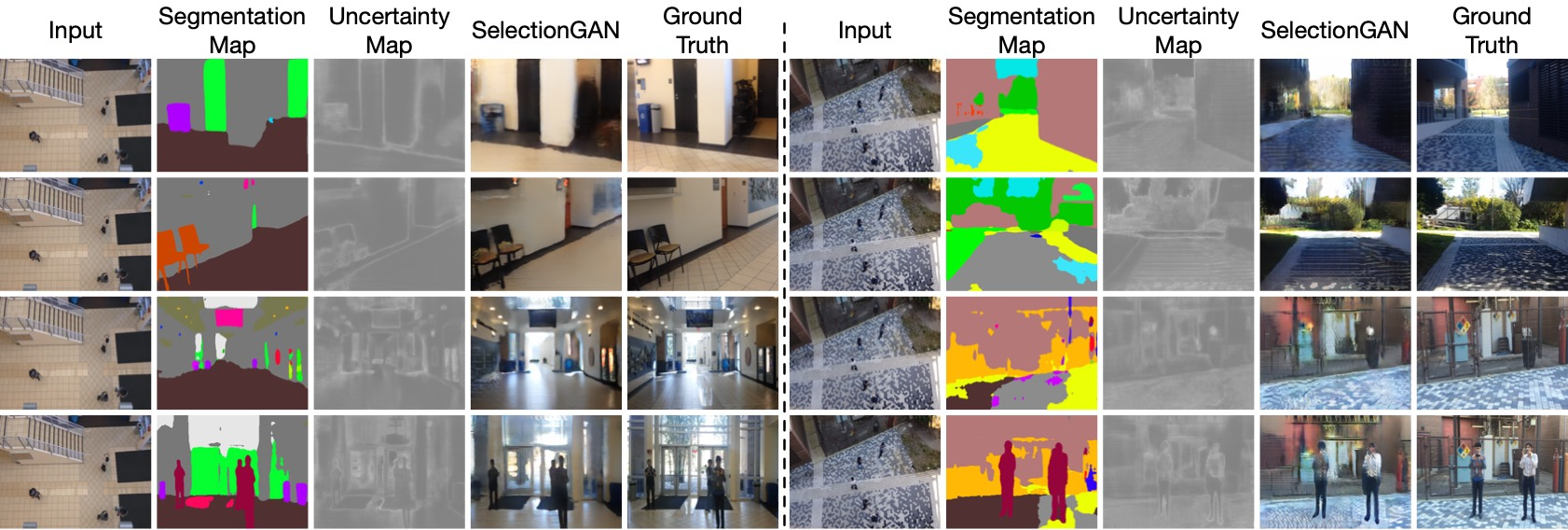}
	\caption{Results of controllable cross-view image translation for both indoor (\textit{left}) and outdoor (\textit{right}) scenes.}
	\label{fig:ego}
	\vspace{-0.2cm}
\end{figure*}

\begin{table*}[!t] \small
	\centering
	\caption{Quantitative results of cross-view image translation on SVA. ($\ast$) Inception Score for real (ground truth) data is 3.1282, 2.4932 and  3.4646 for all, top-1 and top-5 setups, respectively.}
	\resizebox{0.9\linewidth}{!}{% 
		\begin{tabular}{lcccccccccccc} \toprule
			\multirow{2}{*}{Method} & \multicolumn{4}{c}{Accuracy (\%) $\uparrow$}& \multicolumn{3}{c}{Inception Score$^\ast$ $\uparrow$} & \multirow{2}{*}{SSIM $\uparrow$} & \multirow{2}{*}{PSNR $\uparrow$} & \multirow{2}{*}{SD $\uparrow$} & \multirow{2}{*}{KL $\downarrow$} & \multirow{2}{*}{FID $\downarrow$} \\ \cmidrule(lr){2-5} \cmidrule(lr){6-8}
			& \multicolumn{2}{c}{Top-1} & \multicolumn{2}{c}{Top-5} & All & Top-1 & Top-5  \\ \midrule
			X-Pix2pix~\cite{isola2017image}&8.5961 &30.3288 &9.0260   &29.9102  & 2.0131 & 1.7221 & 2.2370 & 0.3206 & 17.9944 & 17.0254 & 19.5533 &859.66 \\
			X-SO~\cite{regmi2019cross}     &7.5146 & 30.9507& 10.3905 & 38.9822 & 2.4951 & 1.8940 & 2.6634 &0.4552  & 21.5312 & 17.5285 & 12.0906 &443.79 \\
			X-Fork~\cite{regmi2018cross}   &17.3794&53.4725 &23.8315  &63.5045  & 2.1888 & 1.9776 & 2.3664 & 0.4235 & 21.2400 & 16.9371 & 4.1925  &129.16\\
			X-Seq~\cite{regmi2018cross}    &19.5056&57.1010 &25.8807  &65.3005  &2.2232  & 1.9842 & 2.4344 & 0.4638 & 22.3411 & 17.4138 & 3.7585  &118.70\\ 
			
			H-Pix2pix~\cite{regmi2019cross}&18.0706& 54.8068& 23.4400 & 62.3072 & 2.1906 & 1.9507 & 2.4069 & 0.4327 & 21.6860 & 16.9468 & 4.2894  &117.13\\
			H-SO~\cite{regmi2019cross}     &5.2444 & 26.4697& 5.2544  & 31.9527 & 2.3202 & 1.9410 & 2.7340 & 0.4457 & 21.7709 & 17.3876 & 12.8761 &1452.88\\
			H-Fork~\cite{regmi2019cross}   &18.0182& 51.0756& 26.6747 & 62.8166 & 2.3202 & 1.9525 & 2.3918 & 0.4240 & 21.6327 & 16.8653 & 4.7246  &109.43\\
			H-Seq~\cite{regmi2019cross}    &20.7391& 57.5378& 28.5517 & 67.4649 & 2.2394 & 1.9892 & 2.4385 & 0.4249 & 21.4770 & 17.5616 & 4.4260  &95.12\\
			H-Regions~\cite{regmi2019cross}&15.4803& 48.0767& 21.8225 & 56.8994 & 2.6328 & 2.0732 & 2.8347 & 0.4044 & 20.9848 & 17.6858 & 6.0638  &88.78\\ 
			Pix2pix++~\cite{isola2017image}&8.8687 & 34.5434& 9.2713  & 35.7490 & 2.5625 & 2.0879 & 2.7961 & 0.3664 & 17.6549 & 18.4015 & 13.1153 &220.23\\
			X-Fork++~\cite{regmi2018cross} &10.2658& 37.8405& 11.4138 & 38.7976 & 2.4280 & 2.0387 & 2.7630 & 0.3406 & 17.3937 & 18.2153 & 10.1403 &166.33\\
			X-Seq++~\cite{regmi2018cross}  &11.2580& 36.8018& 11.9838 & 36.9231 & 2.6849 & 2.1325 & 2.9397 & 0.3617 & 17.4893 & 18.4122 & 11.8560 &154.80\\
			Pix2pixHD \cite{wang2018high} & 35.0018 & 72.9430 & 52.2181&85.6375& 2.5820 & 2.1436 & 2.8730 &0.5437&23.1823&18.9723& 2.6322& 32.79 \\
			GauGAN \cite{park2019semantic} & 34.6740&71.4061&50.1152&81.4900 & 2.6462& \textbf{2.2112} & 2.9550 & 0.5195 & 22.0174 & 18.7762 & 2.6714 & 27.93\\
			SelectionGAN  &33.9055& 71.8779& 50.8878 & 85.0019 & 2.6576 & 2.1279 & 2.9267 & \textbf{0.5752} & \textbf{24.7136} & \textbf{19.7302} & 2.6183  &\textbf{26.09}\\ 	
			SelectionGAN++                 &\textbf{35.9008}& \textbf{73.3249}& \textbf{52.5346} & \textbf{86.9432} & \textbf{2.7370} & 2.1914 & \textbf{3.0271} & 0.5481 & 24.2886 & 19.2001 & \textbf{2.5788}  & 37.17 \\ \bottomrule
		\end{tabular}}
		\label{tab:sva}
		\vspace{-0.2cm}
	\end{table*}

\begin{table}[!t]\small
\centering
\caption{Per-class accuracy and mean IOU for the generated segmentation maps on Dayton.}
	\begin{tabular}{lcc} \toprule
		Method & Per-class Acc. $\uparrow$   & mIOU $\uparrow$   \\ \midrule	
		X-Fork \cite{regmi2018cross} & 0.6262           & 0.4163  \\
		X-Seq \cite{regmi2018cross}  & 0.4783           & 0.3187 \\
		SelectionGAN                 & 0.6415 & 0.5455\\
		SelectionGAN++ & \textbf{0.6619} & \textbf{0.5741} \\
		\bottomrule		
	\end{tabular}
	\label{tab:seg}
	\vspace{-0.4cm}
\end{table}

\noindent \textbf{State-of-the-Art Comparisons.}
We compare our SelectionGAN with several recently proposed state-of-the-art methods, which are Pix2pix~\cite{isola2017image}, Zhai et al.~\cite{zhai2017predicting}, X-Fork~\cite{regmi2018cross}, X-Seq~\cite{regmi2018cross} and X-SO~\cite{regmi2019cross}.
Moreover, to study the effectiveness of SelectionGAN, we introduce five strong baselines which use both segmentation map and RGB image as inputs, including Pix2pix++ \cite{isola2017image}, X-Fork++ \cite{regmi2018cross}, X-Seq++ \cite{regmi2018cross}, Pix2pixHD \cite{wang2018high}, and GauGAN \cite{park2019semantic}.
The comparison results are shown in Table~\ref{tab:cvusa}, \ref{tab:dayton}, and \ref{tab:sva}.
We can observe that SelectionGAN consistently outperforms existing methods on most metrics.
Qualitative results compared with the leading baselines are shown in Fig.~\ref{fig:cvusa_dayton_ego2top} and \ref{fig:dayton}.
We can see that our method generates more clear details on objects/scenes such as road, tress, clouds, car than the other comparison methods. 
Moreover, the results generated by our method are closer to the ground truths in layout and structure.

\noindent \textbf{Visualization of Learned Uncertainty Maps.}
In Fig. \ref{fig:cross_view_coarse_fine} and \ref{fig:ego}, we show some samples of the generated uncertainty maps.
We can see that the generated uncertainty maps learn the layout and structure of the target images.
Note that most textured regions are similar in our generation images, while the junction/edge of different regions is uncertain, and thus the model learns to highlight these parts.

\noindent \textbf{Generated Semantic Guidances.}
Since the proposed methods can reconstruct the semantic guidance (here, the segmentation maps), we also compare the generated semantic guidance with X-Fork~\cite{regmi2018cross} and X-Seq~\cite{regmi2018cross} on Dayton.
Following \cite{regmi2018cross}, we compute the per-class accuracy and mean IOU for the most common classes in this dataset (see Table \ref{tab:seg}).
We see that our SelectionGAN and SelectionGAN++ achieve better results than X-Fork~\cite{regmi2018cross} and X-Seq~\cite{regmi2018cross} on both metrics.
	
\noindent \textbf{Controllable Cross-View Image Translation.}
We further adopt Ego2Top to conduct the controllable cross-view image translation experiments. The results are shown in  Fig.~\ref{fig:ego}. 
Given a single input image and some novel segmentation maps, SelectionGAN is able to generate the same scene but with different viewpoints in both indoor and outdoor environments.

\begin{figure}[!t]\small
	\centering
	\includegraphics[width=0.9\linewidth]{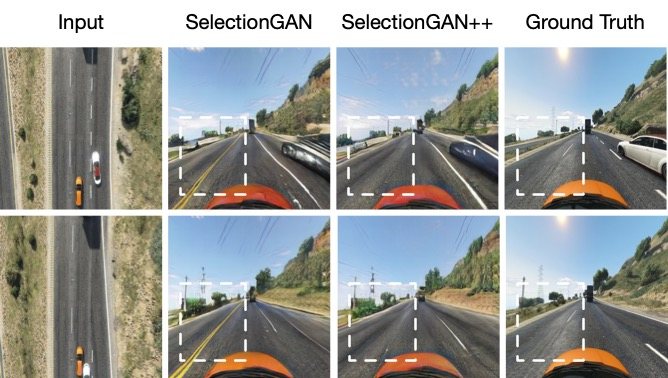}
	\caption{Comparison results of SelectionGAN and SelectionGAN++ on SVA.}
	\label{fig:selectiongan++}
	\vspace{-0.4cm}
\end{figure}

\noindent \textbf{SelectionGAN vs. SelectionGAN++.}
We also provide comparison results of SelectionGAN~\cite{tang2019multi} and SelectionGAN++ in Table \ref{tab:cvusa}, \ref{tab:dayton}, and \ref{tab:sva}.
SelectionGAN++ achieves better results on most metrics, meaning that the proposed multi-scale channel selection module indeed enhances the feature representation, and thus is improving the generation performance.
Note that SelectionGAN++ generates sharper and more realistic images than SelectionGAN, but SelectionGAN has higher pixel-wise similarity scores (i.e., SSIM, PSNR, and SD). This is also observed in other image generation \cite{ma2017pose}, super-resolution \cite{johnson2016perceptual}, and human perceptual judgment \cite{zhang2018unreasonable} tasks. 
From the visualization results in Fig.~\ref{fig:cvusa_dayton_ego2top}, \ref{fig:dayton}, and \ref{fig:selectiongan++}, we see that SelectionGAN++ generates more photo-realistic images with fewer visual artifacts than SelectionGAN on both tasks.
For example, 
SelectionGAN generates road lines in the first and second rows  of Fig.~\ref{fig:selectiongan++}, but there are no road lines in the corresponding ground truths.

\subsection{Results on Facial Expression Generation}
\textbf{Datasets.} We follow C2GAN~\cite{tang2019cycle} and conduct facial expression generation experiments on the Radboud Faces dataset~\cite{langner2010presentation}.
This dataset contains over 8,000 face images with eight different emotional expressions.
We follow C2GAN and all the images are resized to $256 {\times} 256$ without any pre-processing. 
Then, we adopt OpenFace~\cite{amos2016openface} to extract facial landmarks as the semantic guidance.
Consequently, we collect 5,628 training image pairs and 1,407 testing pairs.

\begin{figure*}[!t] \small
	\centering
	\includegraphics[width=0.9\linewidth]{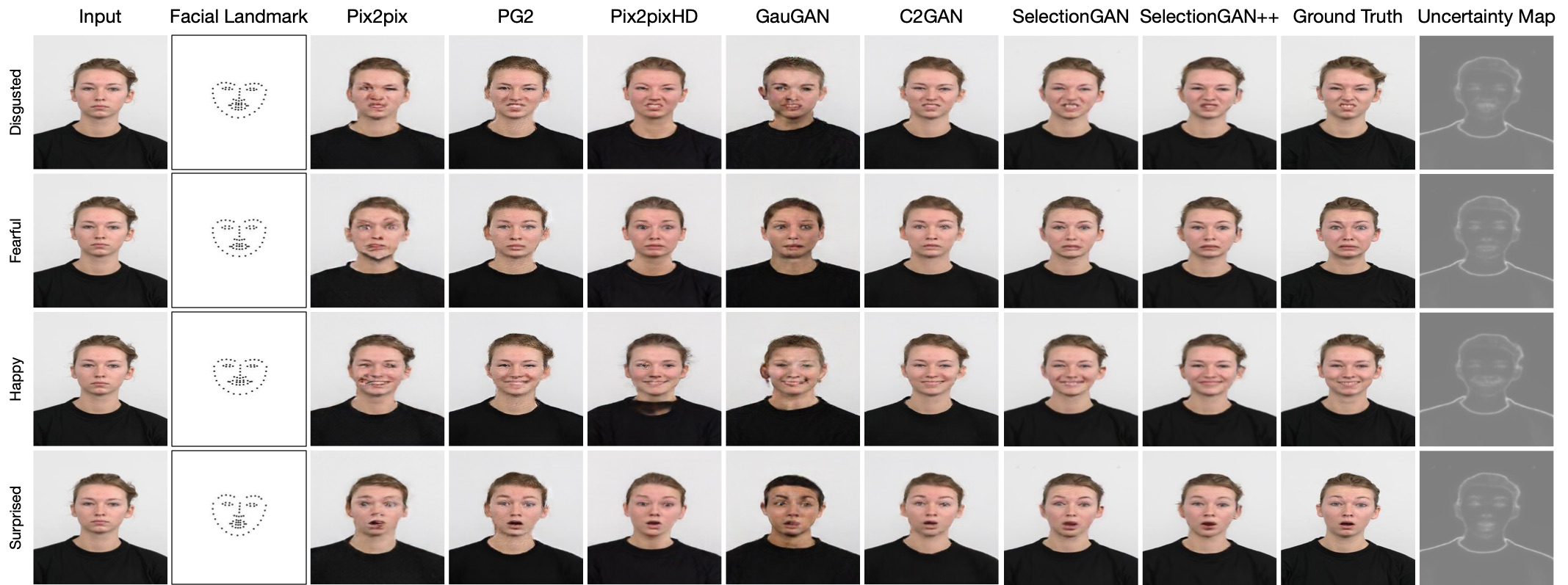}
	\caption{Results of facial expression generation on Radboud Faces.}
	\label{fig:expression_results}
	\vspace{-0.4cm}
\end{figure*}

\noindent \textbf{Parameter Settings.}
Following C2GAN~\cite{tang2019cycle}, the experiments on Radboud Faces are trained for 200 epochs with batch size of 4.

\noindent\textbf{Evaluation Metrics.} 
We follow C2GAN~\cite{tang2019cycle} and employ Structural Similarity (SSIM) \cite{wang2004image} and Peak Signal-to-Noise Ratio (PSNR) to evaluate the quantitative quality of generated images.
Moreover, we adopt Amazon Mechanical Turk (AMT) perceptual studies to evaluate the quality of the generated images.
Specifically, participants were shown a sequence of pairs of images, one a real image and one fake image, and asked to click on the image they thought was real.
The same exact images are presented to the workers for all baselines for fair comparisons.
Finally, we also use a neural network based metric LPIPS~\cite{zhang2018unreasonable} to evaluate the proposed method.

\begin{table}[!t] \small
	\centering
	\caption{Quantitative results of facial expression generation on Radboud Faces.} 
	\resizebox{0.9\linewidth}{!}{% 
		\begin{tabular}{lcccccc} \toprule
			Method                          & AMT $\uparrow$  & SSIM $\uparrow$  & PSNR $\uparrow$   & LPIPS $\downarrow$ \\ \midrule		
			StarGAN~\cite{choi2018stargan}  & 24.7            & 0.8345           & 19.6451           & - \\
			Pix2pix~\cite{isola2017image}  & 13.4            & 0.8217           & 19.9971           & 0.1334 \\ 
			GPGAN~\cite{di2018gp}          & 0.3             & 0.8185           & 18.7211           & 0.2531 \\ 
			PG2~\cite{ma2017pose}          & 28.4            & 0.8462           & 20.1462           & 0.1130 \\  
			Pix2pixHD \cite{wang2018high} & 20.5 & 0.8269 & 24.5621 & 0.1228\\
			GauGAN \cite{park2019semantic} & 10.7 & 0.7528 & 20.8430 & 0.2170 \\
			C2GAN~\cite{tang2019cycle}     & 34.2            & 0.8618           & 21.9192           & 0.0934 \\  
			SelectionGAN                   & 37.5            & 0.8760           & \textbf{27.5671}  & 0.0917 \\  
			SelectionGAN++                & \textbf{39.1}   & \textbf{0.8761}  & 27.5158           & \textbf{0.0905} \\ \bottomrule	
		\end{tabular}}
		\label{tab:result_face}
		\vspace{-0.4cm}
	\end{table}

\begin{figure*}[!t] \small
	\centering
	\includegraphics[width=0.9\linewidth]{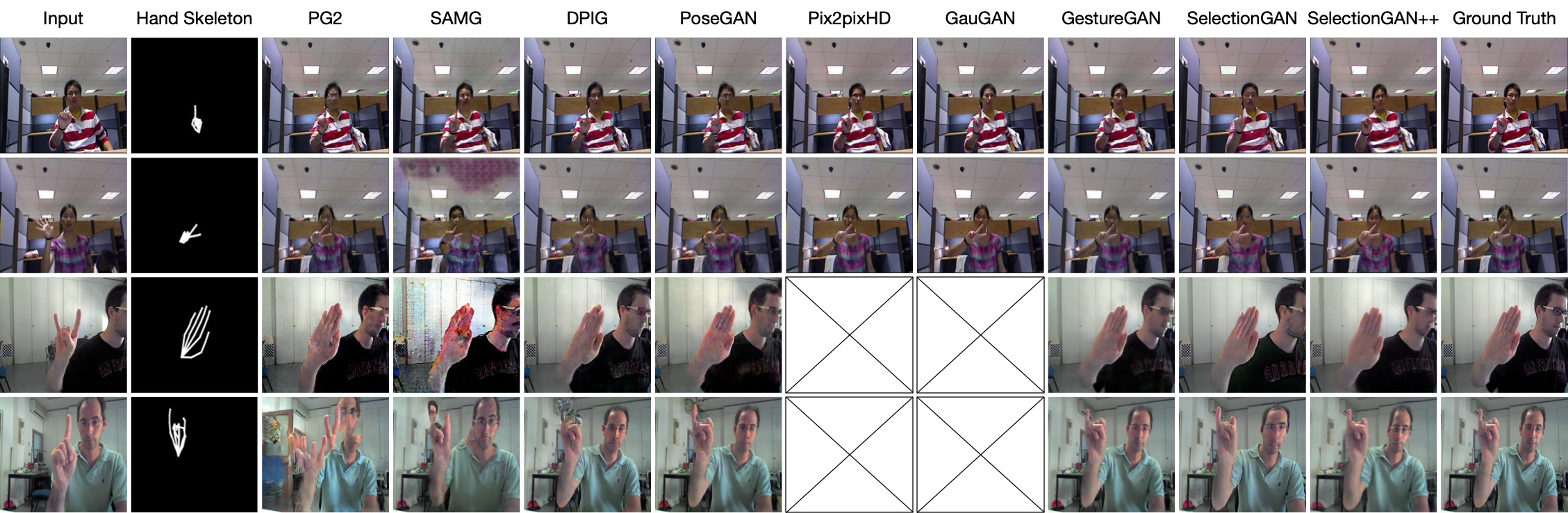}
	\caption{Results of hand gesture-to-gesture translation on NTU Hand Digit (\textit{top two rows}) and Senz3D (\textit{bottom two rows}).}
	\label{fig:gesture_comp_results}
	\vspace{-0.2cm}
\end{figure*}

\begin{table*}[!t] \small
	\centering
	\caption{Quantitative results of hand gesture-to-gesture translation on NTU Hand Digit and Senz3D.}
		\resizebox{0.9\linewidth}{!}{% 
		\begin{tabular}{lcccccccccc} \toprule
			\multirow{2}{*}{Method} & \multicolumn{5}{c}{NTU Hand Digit} & \multicolumn{5}{c}{Senz3D} \\ \cmidrule(lr){2-6} \cmidrule(lr){7-11} 	
			& PSNR $\uparrow$ & IS $\uparrow$  & AMT $\uparrow$  & FID $\downarrow$ & FRD $\downarrow$ & PSNR $\uparrow$ & IS $\uparrow$  & AMT $\uparrow$  & FID $\downarrow$ & FRD $\downarrow$ \\ \midrule
			PG2~\cite{ma2017pose}                & 28.2403 & 2.4152          & 3.5   & 24.2093 & 2.6319 & 26.5138 & 3.3699 & 2.8 & 31.7333 & 3.0933 \\ 
			SAMG~\cite{yan2017skeleton}          & 28.0185 & 2.4919          & 2.6   & 31.2841 & 2.7453 & 26.9545 & 3.3285 & 2.3 & 38.1758 & 3.1006\\ 
			DPIG~\cite{ma2017disentangled}       & 30.6487 & 2.4547          & 7.1   & \textbf{6.7661}  & 2.6184 & 26.9451 & 3.3874 & 6.9 & 26.2713 & 3.0846\\  
			PoseGAN~\cite{siarohin2018deformable} & 29.5471 & 2.4017          & 9.3   & 9.6725  & 2.5846 & 27.3014 & 3.2147 & 8.6 & 24.6712 & 3.0467 \\ 
			Pix2pixHD \cite{wang2018high} & \textbf{38.1295} & 2.2358& 21.3 & 8.4003& \textbf{1.1475} & - & - & - & - & -\\
			GauGAN \cite{park2019semantic} & 32.2218 & \textbf{2.6210} & 13.2 &18.4373& 1.8229 & - & - & - & - & - \\
			GestureGAN~\cite{tang2018gesturegan}  & 32.6091 &2.5532 & \textbf{26.1}  & 7.5860  & 2.5223 & 27.9749 & \textbf{3.4107} & \textbf{22.6}& \textbf{18.4595} & 2.9836\\
			SelectionGAN                         & 30.6465 & 2.4472          & 15.8  & 16.2159 & 2.1560 & 30.4036 & 2.4595 & 14.1& 30.9775 & 2.7014 \\
			SelectionGAN++ & 31.4580 & 2.5197 & 20.9 & 12.4843 & 2.0221 & \textbf{31.1875} & 2.8194 & 18.7 & 23.6390 & \textbf{2.6711} \\
			\bottomrule		
		\end{tabular}}
		\label{tab:gesture_comp}
		\vspace{-0.4cm}
	\end{table*}

\noindent\textbf{State-of-the-Art Comparisons.}
We compare the proposed SelectionGAN with several state-of-the-art methods, i.e., StarGAN~\cite{choi2018stargan}, Pix2pix~\cite{isola2017image}, GPGAN~\cite{di2018gp}, PG2~\cite{ma2017pose}, Pix2pixHD \cite{wang2018high}, GauGAN \cite{park2019semantic}, and C2GAN~\cite{tang2019cycle}.
Quantitative results of the SSIM, PSNR, LPIPS, and  AMT metrics are show in Table~\ref{tab:result_face}.
We can see that the proposed SelectionGAN and SelectionGAN++ achieve better results than the existing methods on all metrics, validating the effectiveness of our methods.
Note that GauGAN achieves unsatisfactory results in this task since it is proposed to use segmentation maps as input.
However, this task uses facial landmarks as guidances, which is quite different from segmentation maps.
On the contrary, our methods achieve good results in this task, which further proves the generalizability of our proposed methods.
Qualitative results are shown in Fig.~\ref{fig:expression_results}.
Clearly, the image generated by our SelectionGAN and SelectionGAN++ are more sharper and contains more image details compared to other leading methods.

\noindent\textbf{Visualization of Learned Uncertainty Maps.} 
We also show the learned uncertainty maps in Fig.~\ref{fig:expression_results}.
We observe that the proposed SelectionGAN can generate different uncertainty maps according to different facial expressions, which means the proposed model can learn the difference between different expression domains.

\noindent\textbf{Efficiency.}
We also compared the proposed methods with existing methods on facial expression generation.
Our proposed SelectionGAN and SelectionGAN++ takes about 24
and 27 hours to finish the training on a single
NVIDIA DGX1 V100 GPU, while C2GAN, GauGAN, Pix2pixHD, and PG2 takes
around 27, 32, 36, and 30 hours, respectively.
This also validates that the proposed methods are efficient.

\noindent \textbf{SelectionGAN vs. SelectionGAN++.}
We also provide comparison results of SelectionGAN~\cite{tang2019multi} and SelectionGAN++ in Table \ref{tab:result_face}.
SelectionGAN++ achieves better results than SelectionGAN on most metrics, i.e., AMT, SSIM, and LPIPS.
Meanwhile, SelectionGAN++ generates more realistic details (e.g., eyes and month) than SelectionGAN (see Fig.~\ref{fig:expression_results}).

\subsection{Results on Hand Gesture Translation}
\noindent\textbf{Datasets.} We follow GestureGAN~\cite{tang2018gesturegan} and conduct experiments on both NTU Hand Digit~\cite{ren2013robust} and Senz3D~\cite{memo2018head} datasets.
NTU Hand Digit dataset contains 75,036 and 9,600 image pairs for training and testing sets, each of which is comprised of two
images of the same person but different gestures. 
For Senz3D, which contains 135,504 pairs and 12,800 pairs for training and testing.

\noindent \textbf{Parameter Settings.}
Images on both datasets are resized to $256{\times}256$, and we enabled image flipping and random crops for data augmentation.
Following GestureGAN~\cite{tang2018gesturegan}, the experiments on both datasets are trained for 20 epochs with batch size of 4.

\begin{figure*}[!t] \small
	\centering
	\includegraphics[width=1\linewidth]{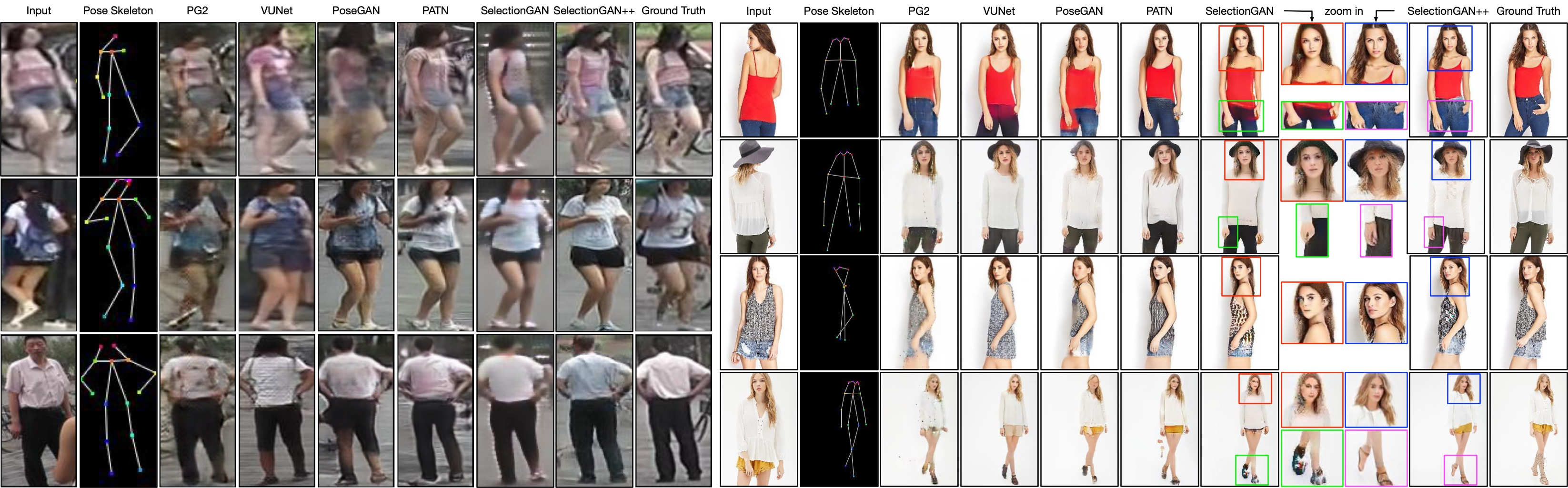}
	\caption{Results of person image generation on Market-1501 (\textit{left}) and DeepFashion (\textit{right}). Key differences in DeepFashion are highlighted by colored boxes.}
	\label{fig:person_reuslts}
	\vspace{-0.2cm}
\end{figure*}

\begin{table*}[!t]\small
	\centering
	\caption{(\textit{left}) Quantitative results of person image generation on Market-1501 and DeepFashion. ($\ast$) denotes the results tested on our test set. (\textit{right}) User study (\%) of person image generation. For each comparison, the participant is asked to answer two questions, i.e., `Q1: Which generated image is more realistic regardless of the target image?', and `Q2: Which generated image matches the conditioning image better (e.g., clothes)?'.
	}
	\resizebox{0.56\linewidth}{!}{
	\begin{tabular}{lccccccc} \toprule
		\multirow{2}{*}{Method} & \multicolumn{4}{c}{Market-1501} & \multicolumn{2}{c}{DeepFashion} \\ \cmidrule(lr){2-5} \cmidrule(lr){6-7} 
		& SSIM $\uparrow$ & IS $\uparrow$   & M-SSIM $\uparrow$& M-IS $\uparrow$ & SSIM $\uparrow$ & IS $\uparrow$    \\ \midrule
		PG2~\cite{ma2017pose}                          & 0.253 & 3.460 & 0.792     & 3.435   & 0.762 & 3.090 \\
		DPIG~\cite{ma2017disentangled}                 & 0.099 & 3.483 & 0.614     & 3.491   & 0.614 & 3.228 \\
		PoseGAN~\cite{siarohin2018deformable}          & 0.290 & 3.185 & 0.805     & 3.502   & 0.756 & 3.439 \\ 
		C2GAN~\cite{tang2019cycle}                     & 0.282 & 3.349 & 0.811     & 3.510   & -   & -  \\ 
		BTF~\cite{albahar2019guided}                   & -  & -   & -       & -     & 0.767 & 3.220 \\ \midrule
		PG2$^\ast$~\cite{ma2017pose}                    & 0.261 & 3.495 & 0.782     & 3.367   & 0.773 & 3.163 \\ 
		PoseGAN$^\ast$~\cite{siarohin2018deformable}    & 0.291 & 3.230 & 0.807     & 3.502   & 0.760 & 3.362\\ 
		Pix2pixHD$^\ast$ \cite{wang2018high}  & - & - & - & - & 0.762 & 3.224\\
		GauGAN$^\ast$ \cite{park2019semantic} & - & - & - & - & 0.754 & 3.165 \\
		VUNet$^\ast$~\cite{esser2018variational}        & 0.266 & 2.965 & 0.793     & 3.549   & 0.763 & 3.440 \\
		PATN$^\ast$~\cite{zhu2019progressive}  & 0.311 & 3.323 & 0.811     & \textbf{3.773}   & 0.773 & 3.209 \\  
		SelectionGAN                                   & 0.331 & 3.449 & 0.816 & 3.376   & 0.776 & 3.341 \\ 
		SelectionGAN++ & \textbf{0.333} & \textbf{3.512} & \textbf{0.818} & 3.651 & \textbf{0.778} & \textbf{3.445} \\ \midrule	
		Real Data                                      & 1.000 & 3.890 & 1.000     & 3.706   & 1.000 & 4.053 \\	
		\bottomrule	
	\end{tabular}}
        \begin{tabular}{lccccccc} \toprule
			\multirow{2}{*}{Method}  & \multicolumn{2}{c}{Market-1501} & \multicolumn{2}{c}{DeepFashion} \\ \cmidrule(lr){2-3} \cmidrule(lr){4-5} 
			& Q1 $\uparrow$ & Q2 $\uparrow$ & Q1 $\uparrow$ & Q2 $\uparrow$    \\ \midrule	
			PG2~\cite{ma2017pose}                        & 4.2  & 3.1    & 6.3   & 6.5 \\
			PoseGAN~\cite{siarohin2018deformable}        & 8.3 & 6.7     & 10.5 & 8.2 \\ 
			C2GAN~\cite{tang2019cycle}                   & 16.1 & 17.6   & -   & -     \\
			PATN~\cite{zhu2019progressive}               & 20.3 & 19.9   & 22.9 & 23.1 \\  
			SelectionGAN                                & 23.7 & 24.2   & 28.1 & 29.3 \\
			SelectionGAN++                              & \textbf{27.4} & \textbf{28.5} & \textbf{32.2} & \textbf{32.9} \\
			\bottomrule	
		\end{tabular}
	\label{tab:pose_reuslts}
	\vspace{-0.4cm}
\end{table*}

\noindent\textbf{Evaluation Metrics.}
We follow~\cite{tang2018gesturegan} and employ Peak Signal-to-Noise Ratio (PSNR), Inception score (IS)~\cite{salimans2016improved}, Fr\'echet Inception Distance (FID)~\cite{heusel2017gans}, and Fr\'echet ResNet Distance (FRD)~\cite{tang2018gesturegan} to evaluate the generated images.
Moreover, we follow the same settings as in~\cite{isola2017image,tang2018gesturegan} to conduct the Amazon Mechanical Turk (AMT) perceptual studies.

\noindent\textbf{State-of-the-Art Comparisons.}
We compare the proposed methods with the leading hand gesture translation methods, i.e., PG2~\cite{ma2017pose}, SAMG~\cite{yan2017skeleton}, DPIG~\cite{ma2017disentangled}, PoseGAN~\cite{siarohin2018deformable}, Pix2pixHD \cite{wang2018high}, GauGAN \cite{park2019semantic}, and GestureGAN~\cite{tang2018gesturegan}.
Comparison results are shown in Table~\ref{tab:gesture_comp}.
We can see that our SelectionGAN and SelectionGAN++ achieve competitive results on both datasets.
Note that GestureGAN achieves better results than
the proposed methods. The reason is that GestureGAN is
carefully tailored and designed for this task, meaning that GestureGAN is fine-turned to this task with the network structure, loss objective, and hyper-parameter selection.  
However, the proposed methods are novel and unified GAN models, which can be used to handle various settings of guided image-to-image translation without modifying the network structure, the loss objective, and hyper-parameters.
Qualitative results compared with existing methods are shown in Fig.~\ref{fig:gesture_comp_results}. 
We can see that our SelectionGAN and SelectionGAN++ also generate photo-realistic results on this challenging task.
Moreover, we show the learned uncertainty maps in Fig.~\ref{fig:gesture_results}.

\noindent \textbf{Controllable Hand Gesture Translation.}
In Fig.~\ref{fig:gesture_results}, we provide results of controllable hand gesture translation.
We can see that the proposed SelectionGAN can translate a single input image into several output images while each one respecting the constraints specified in the provided hand skeleton.

\noindent \textbf{SelectionGAN vs. SelectionGAN++.}
We also provide comparison results of SelectionGAN~\cite{tang2019multi} and SelectionGAN++. The results of hand gesture translation are shown in Table \ref{tab:gesture_comp}.
We can see that SelectionGAN++ achieves better results than SelectionGAN on all metrics.
Meanwhile, SelectionGAN++ generates more photo-realistic results than SelectionGAN, as shown in Fig.~\ref{fig:gesture_results}.

\subsection{Results on Person Image Generation}
\textbf{Datasets.} We follow PATN~\cite{zhu2019progressive} and conduct person image generation experiments on both Market-1501~\cite{zheng2015scalable} and DeepFashion~\cite{liu2016deepfashion} datasets.
Following~\cite{zhu2019progressive}, we collect 263,632 and 12,000 pairs for training and testing on Market-1501. 
For DeepFashion, 101,966 and 8,570 pairs are randomly selected for training and testing. 

\noindent \textbf{Parameter Settings.}
Following PATN~\cite{zhu2019progressive}, images are re-scaled to $128 {\times} 64$ and $256 {\times} 256$ on Market-1501 and DeepFashion datasets, respectively. 
Moreover, the experiments on both datasets are trained for around 90k iteration with batch size of 32 and 12 on Market-1501 and DeepFashion, respectively.

\noindent\textbf{Evaluation Metrics.} 
We follow previous works~\cite{ma2017pose, siarohin2018deformable, tang2019cycle, siarohin2018deformable, zhu2019progressive} and adopt Structure Similarity (SSIM)~\cite{wang2004image}, Inception score (IS)~\cite{salimans2016improved} and their corresponding masked versions, i.e., M-SSIM
and M-IS, as our evaluation metrics.
We also recruit 30 volunteers to conduct a user study. Specifically, given six results (four generated by existing methods, two generated by our proposed SelectionGAN and SelectionGAN++), each participant needs to answer two questions: `Q1: Which generated image is more realistic regardless of the target image?' and `Q2: Which generated image matches the conditioning image better (e.g., clothes)?'.

\noindent\textbf{State-of-the-Art Comparisons.}
We compare the proposed SelectionGAN and SelectionGAN++ with several leading person image generation methods, i.e., PG2~\cite{ma2017pose}, DPIG~\cite{ma2017disentangled}, PoseGAN~\cite{siarohin2018deformable}, VUNet~\cite{esser2018variational}, C2GAN~\cite{tang2019cycle}, BTF~\cite{albahar2019guided}, Pix2pixHD \cite{wang2018high}, GauGAN \cite{park2019semantic},  and PATN~\cite{zhu2019progressive}.
Quantitative results of the SSIM, IS, M-SSIM, and M-IS metrics are show in Table~\ref{tab:pose_reuslts}(\textit{left}).
We see that the proposed SelectionGAN and SelectionGAN++ achieve competitive performance compared with the carefully designed methods on this task such as PATN~\cite{zhu2019progressive} and PoseGAN~\cite{siarohin2018deformable}.
Moreover, we show the user study results in Table~\ref{tab:pose_reuslts}(\textit{right}).
We observe that our methods achieve better results over~\cite{siarohin2018deformable, tang2019cycle, ma2017pose, zhu2019progressive} in terms of both image realism and style consistency, further validating that our generated images are more photo-realistic.
Qualitative results are shown in Fig.~\ref{fig:person_reuslts}.
The images generated by our SelectionGAN and SelectionGAN++ are more realistic and sharp compared with other leading methods.
Moreover, the person layouts of the generated images by our methods are closer to the target skeletons.

\begin{figure*}[!t] \small
	\centering
	\includegraphics[width=0.85\linewidth]{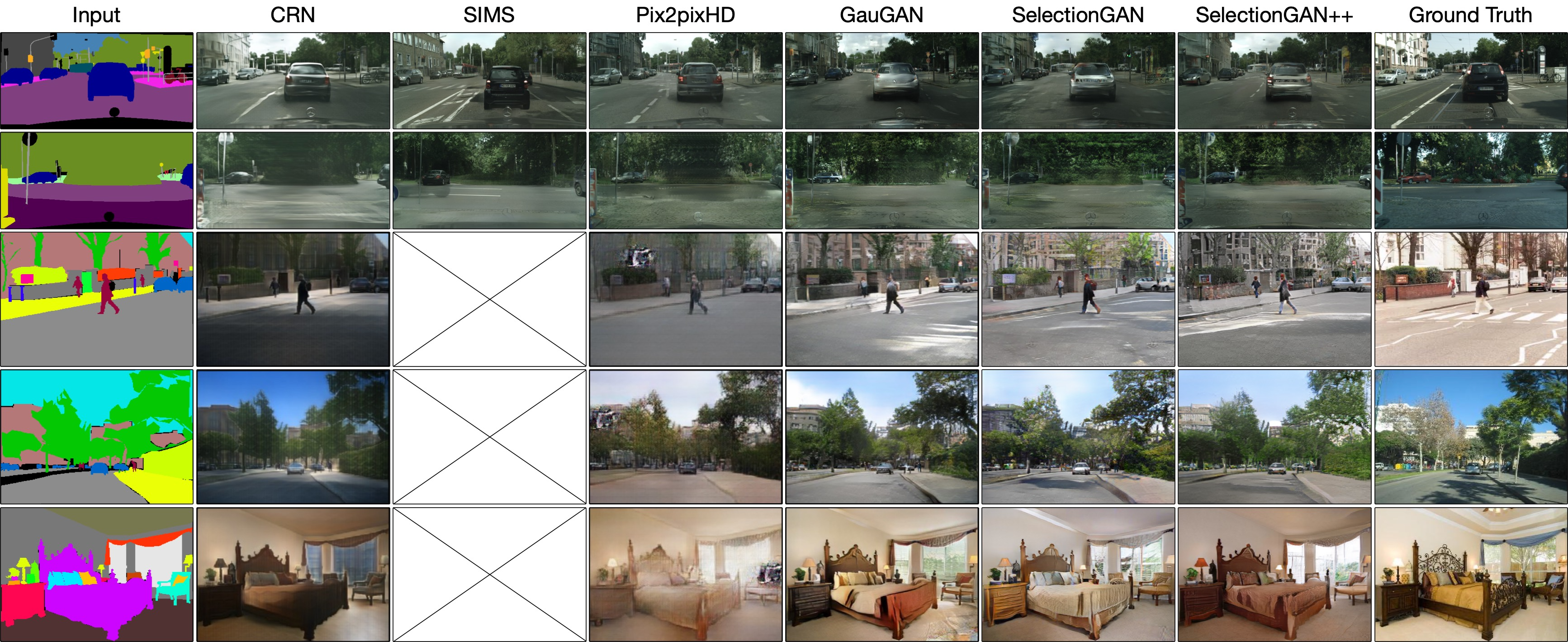}
	\caption{Results of semantic image synthesis on Cityscapes (\textit{top two rows}) and ADE20K (\textit{bottom three rows}).}
	\label{fig:cityscapes_ade}
	\vspace{-0.2cm}
\end{figure*}

\begin{table*}[!t]\small
	\centering
	\caption{(\textit{left}) Quantitative results of semantic image synthesis on Cityscapes and ADE20K. (\textit{right}) User preference study of semantic image synthesis on Cityscapes and ADE20K. The numbers indicate the percentage of users who favor the results of the proposed SelectionGAN++ over the competing method. 
	}
	\resizebox{0.56\linewidth}{!}{
	\begin{tabular}{lcccccc} \toprule
			\multirow{2}{*}{Method}   & \multicolumn{3}{c}{Cityscapes} & \multicolumn{3}{c}{ADE20K} \\ \cmidrule(lr){2-4} \cmidrule(lr){5-7} 
			& mIoU $\uparrow$ & Acc $\uparrow$  & FID  $\downarrow$ & mIoU $\uparrow$    & Acc $\uparrow$  & FID  $\downarrow$  \\ \midrule
			CRN~\cite{chen2017photographic}  & 52.4  & 77.1 & 104.7  & 22.4 & 68.8 & 73.3 \\
			SIMS~\cite{qi2018semi}           & 47.2  & 75.5 & \textbf{49.7} & - & - & - \\
			Pix2pixHD~\cite{wang2018high}    & 58.3  & 81.4 & 95.0  & 20.3 & 69.2 & 81.8 \\
			GauGAN~\cite{park2019semantic}   & 62.3  & 81.9 & 71.8  & 38.5 & 79.9 & 33.9 \\
			SelectionGAN                     & 63.8  & 82.4 & 65.2  & 40.1 & 81.2 & 33.1 \\
			SelectionGAN++ & \textbf{64.5} & \textbf{82.7} & 63.4 & \textbf{41.7} & \textbf{81.5} & \textbf{32.2} \\ \bottomrule
		\end{tabular}}
        \resizebox{0.42\linewidth}{!}{
        	\begin{tabular}{lcc} \toprule
			AMT $\uparrow$                           & Cityscapes  & ADE20K  \\ \midrule
			Ours vs. CRN~\cite{chen2017photographic} & 65.80       & 72.15 \\
			Ours vs. Pix2pixHD~\cite{wang2018high}   & 60.93       & 80.61 \\ 
			Ours vs. SIMS~\cite{qi2018semi}          & 56.78       & -   \\
			Ours vs. GauGAN~\cite{park2019semantic}  & 55.22       & 57.54 \\ 
			Ours vs. SelectionGAN & 53.17 & 55.75 \\ \bottomrule
		\end{tabular}}
	\label{tab:semantic}
	\vspace{-0.3cm}
\end{table*}

\noindent \textbf{SelectionGAN vs. SelectionGAN++.}
We also provide comparison results of SelectionGAN~\cite{tang2019multi} and SelectionGAN++ in Table~\ref{tab:pose_reuslts}.
We see that SelectionGAN++ achieves better results than SelectionGAN on all metrics.
Moreover, the results of the user study indicate that SelectionGAN++ generates much better results than SelectionGAN. 
Meanwhile, SelectionGAN++ generates more photo-realistic results (especially in DeepFashion) than SelectionGAN (see Fig.~\ref{fig:person_reuslts}).
In order to better prove that SelectionGAN++ produces more realistic images than SelectionGAN, we provide the comparison in a zoomed-in manner on the DeepFashion dataset in Fig.~\ref{fig:person_reuslts}.
For example, in the last row, SelectionGAN++ generates better hair, face, and feet than SelectionGAN.

\subsection{Results on Semantic Image Synthesis}
To explore the generality of SelectionGAN and SelectionGAN++ on other generation tasks, we also conduct experiments on the semantic image synthesis task.
Specifically, we adopt GauGAN \cite{park2019semantic} as our backbone network in this task and we combine it with the proposed multi-channel attention selection module to form our final model.

\noindent\textbf{Datasets.} We follow GauGAN~\cite{park2019semantic} and conduct semantic image synthesis experiments on two challenging datasets, i.e., Cityscapes~\cite{cordts2016cityscapes} and ADE20K~\cite{zhou2017scene}.
The training and testing set sizes of Cityscapes are 2,975 and 500, respectively. 
For ADE20K, which contains 150 semantic classes, and has 20,210 training and 2,000 validation images.

\noindent \textbf{Parameter Settings.}
Images are re-scaled to $512 {\times} 256$ and $256 {\times} 256$ on Cityscapes and ADE20K datasets, respectively.
Following GauGAN~\cite{park2019semantic}, the experiments on both datasets are trained for 200 epochs with batch size of 32.

\noindent\textbf{Evaluation Metrics.} We Follow~\cite{park2019semantic} and employ the mean
Intersection-over-Union (mIoU) and pixel accuracy (Acc) to measure the segmentation accuracy. 
Specifically, we adopt the state-of-the-art segmentation networks to evaluate the generated images, i.e., DRN-D-105~\cite{yu2017dilated} for Cityscapes and UperNet101~\cite{xiao2018unified} for ADE20K.
We also employ the Fr\'echet Inception Distance (FID)~\cite{heusel2017gans} to measure the distance between
the distribution of generated samples and the distribution of real samples. 
Finally, we follow GauGAN and employ Amazon Mechanical Turk (AMT) to measure the perceived visual fidelity of the generated images.

\noindent\textbf{State-of-the-Art Comparisons.}
We adopt several leading semantic image synthesis methods as our baselines, i.e., Pix2pixHD~\cite{wang2018high}, CRN~\cite{chen2017photographic}, SIMS~\cite{qi2018semi}, and GauGAN~\cite{park2019semantic}.
The results of mIoU, Acc, and FID are show in Table~\ref{tab:semantic}(\textit{left}).
We note that our SelectionGAN and SelectionGAN++ achieve better results than the existing competing methods on both mIoU and Acc metrics.
For FID, SelectionGAN and SelectionGAN++ are only worse than SIMS on Cityscapes.
However, SIMS has poor segmentation results. 
Moreover, we follow GauGAN and provide AMT results in Table~\ref{tab:semantic}(\textit{right}).
We observe that users favor our translated images on both datasets compared with existing leading methods.
Qualitative results compared with the exiting methods are shown in Fig.~\ref{fig:cityscapes_ade}. We observe that SelectionGAN and SelectionGAN++ produce much better results with fewer visual artifacts than exiting methods.

\begin{figure}[!t] \small
	\centering
	\includegraphics[width=1\linewidth]{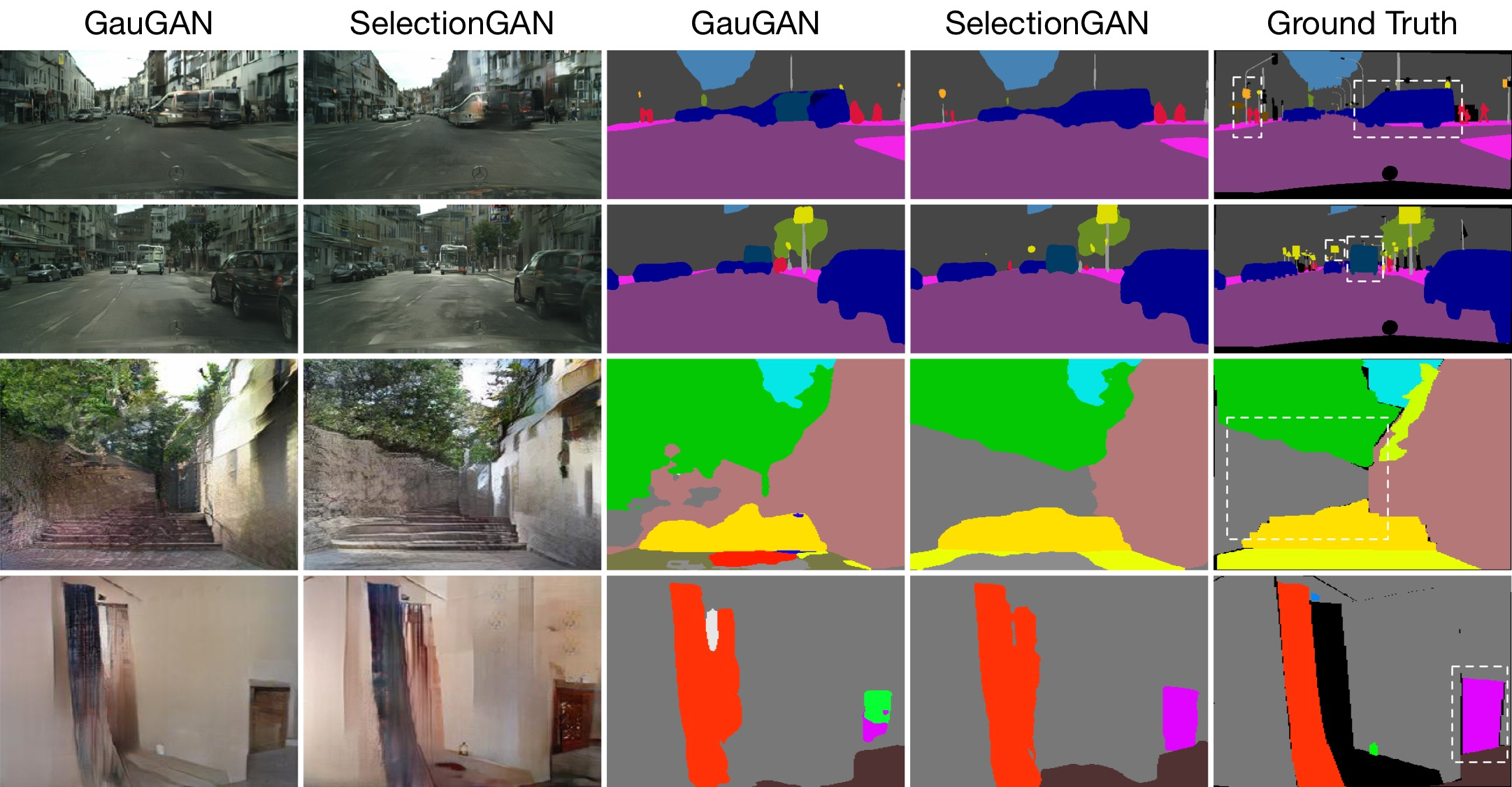}
	\caption{Generated segmentation maps on Cityscapes (\textit{top two rows}) and ADE20K (\textit{bottom two rows}).}
	\label{fig:cityscapes_ade_seg}
	\vspace{-0.4cm}
\end{figure}

\noindent\textbf{Visualization of Generated Segmentation Maps.}
We follow GauGAN and apply pre-trained segmentation networks
on the generated images to produce segmentation maps. 
The intuition behind this is that if the generated images are realistic, a well-trained semantic segmentation model should be able to predict the ground truth label.
The results compared with the state-of-the-art method GauGAN are shown in Fig.~\ref{fig:cityscapes_ade_seg}.
We observe that the proposed SelectionGAN generates better semantic maps than GauGAN on both datasets.

\noindent \textbf{SelectionGAN vs. SelectionGAN++.}
We also provide comparison results of SelectionGAN~\cite{tang2019multi} and SelectionGAN++ in Table~\ref{tab:semantic}.
SelectionGAN++ achieves better results than SelectionGAN on all metrics, i.e., mIoU, Acc, and FID.
Moreover, the results of the user study indicate that SelectionGAN++ generates more photo-realistic results than SelectionGAN. 
We also note that SelectionGAN++ generates better results than SelectionGAN on both datasets (see Fig.~\ref{fig:cityscapes_ade}).
\section{Conclusion}
\label{sec:con}
We propose SelectionGAN to address a novel image synthesis task by conditioning on an input image and several conditional semantic guidances. In particular, we adopt a cascade strategy to divide the generation procedure into two stages. Stage I aims to capture the semantic structure of the target image and Stage II focuses on more appearance details via the proposed multi-scale spatial pooling \& channel selection and the multi-channel attention selection modules. We also propose an uncertainty map guided pixel loss to solve the inaccurate semantic guidance issue for better optimization. Extensive experimental results on four guided image-to-image translation and semantic image synthesis tasks with 11 public datasets show that our method obtains much better results than the state-of-the-art models. 

\noindent \textbf{Acknowledgments.}
This work has been supported by the Italy-China collaboration project TALENT, by the EU H2020 project AI4Media (No. 951911) and by the PRIN project PREVUE  (Prot. 2017N2RK7K).

%\clearpage
\small
\bibliographystyle{IEEEtran}
\bibliography{reference}

\begin{IEEEbiography}[{\includegraphics[width=1in,height=1.25in,clip,keepaspectratio]{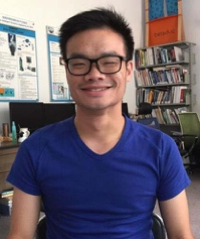}}]{Hao Tang}
is currently a Postdoctoral with Computer Vision Lab, ETH Zurich, Switzerland.
He received the master’s degree from the School of Electronics and Computer Engineering, Peking University, China and the Ph.D. degree from the Multimedia and Human Understanding Group, University of Trento, Italy.
He was a visiting scholar in the Department of Engineering Science at the University of Oxford. His research interests are deep learning, machine learning, and their applications to computer vision.
\end{IEEEbiography}

\begin{IEEEbiography}[{\includegraphics[width=1in,height=1.25in,clip,keepaspectratio]{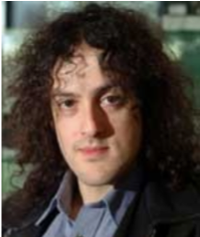}}]{Philip H. S. Torr} received the PhD degree from
	Oxford University. After working for another three
	years at Oxford, he worked for six years for
	Microsoft Research, first in Redmond, then in
	Cambridge, founding the vision side of the Machine Learning and Perception Group. He is now
	a professor at Oxford University. He has won
	awards from top vision conferences, including
	ICCV, CVPR, ECCV, NIPS and BMVC. He is a
	senior member of the IEEE and a Royal Society
	Wolfson Research Merit Award holder.
\end{IEEEbiography}

\begin{IEEEbiography}[{\includegraphics[width=1in,height=1.25in,clip,keepaspectratio]{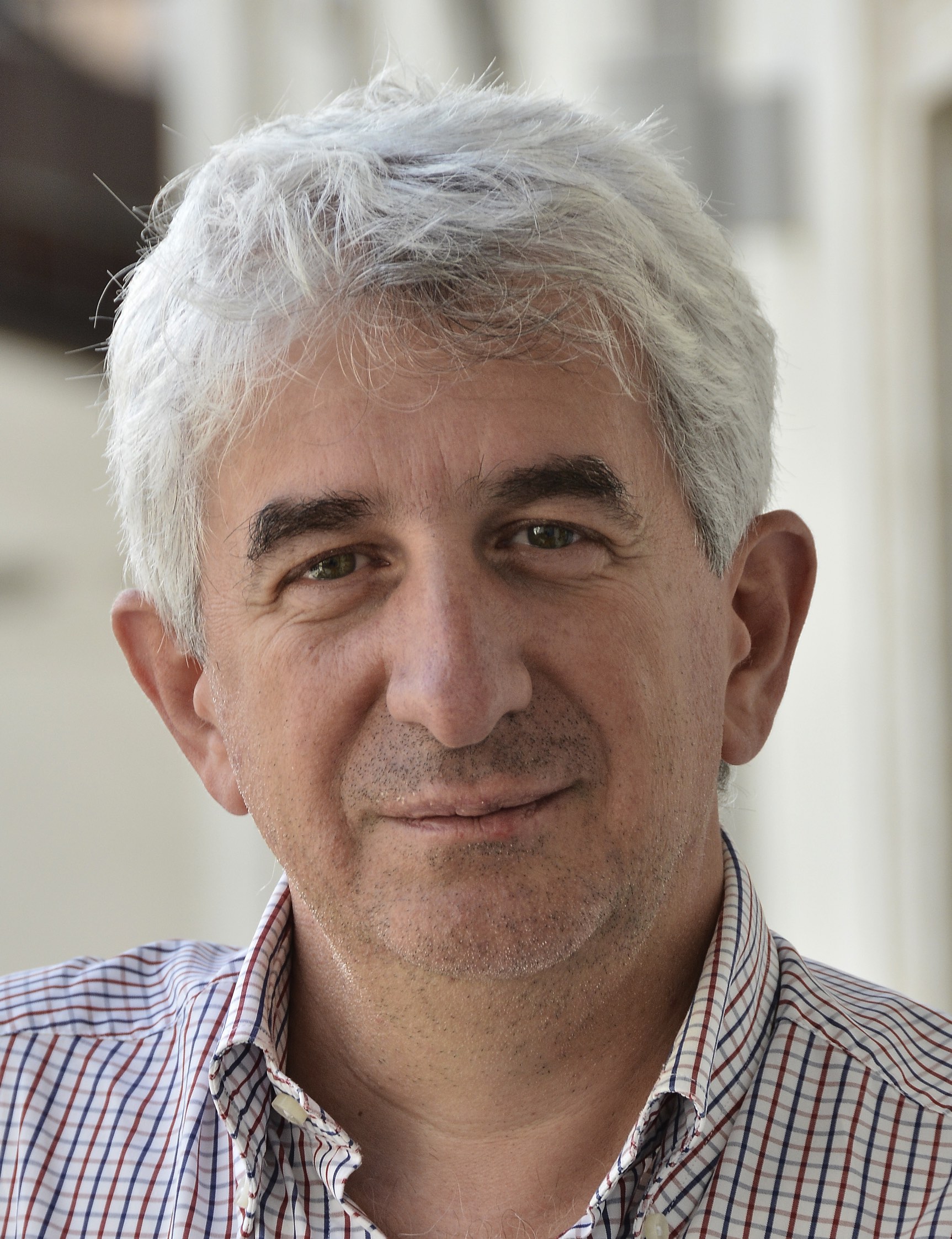}}]{Nicu Sebe} 
is Professor in the University of Trento, Italy, where he is leading the research in the areas of multimedia analysis and human behavior understanding. He was the General Co-Chair of the IEEE FG 2008 and ACM Multimedia 2013.  He was a program chair of ACM Multimedia 2011 and 2007, ECCV 2016, ICCV 2017 and ICPR 2020.  He is a general chair of ACM Multimedia 2022 and a program chair of ECCV 2024. He is a fellow of IAPR.
\end{IEEEbiography}

% You can push biographies down or up by placing
% a \vfill before or after them. The appropriate
% use of \vfill depends on what kind of text is
% on the last page and whether or not the columns
% are being equalized.

%\vfill

% Can be used to pull up biographies so that the bottom of the last one
% is flush with the other column.
%\enlargethispage{-5in}

% that's all folks
\end{document}